\documentclass[10pt,journal,compsoc,twoside]{IEEEtran}
\usepackage{booktabs} % For formal tables

\usepackage{subfigure}
\usepackage{amssymb}
\usepackage{amsmath,bm}
\usepackage{color}
\usepackage{url}
\usepackage{algorithm}
\usepackage{algorithmic}
\usepackage{mdwlist}  %itemize used in the algorithm
\usepackage{multirow}

\newtheorem{definition}{Definition}
\newtheorem{insight}{Insight}

\newcommand{\fs}{{\sf fs}}
\newcommand{\bs}{{\sf bs}}

\newcommand{\tabincell}[2]{\begin{tabular}{@{}#1@{}}#2\end{tabular}}
\newcommand{\ignore}[1]{}

\PassOptionsToPackage{hyphens}{url}
\usepackage[breaklinks,colorlinks,bookmarks=false]{hyperref}
\hypersetup{citecolor=black,linkcolor=black,urlcolor=black}

\ifCLASSINFOpdf
   \usepackage[pdftex]{graphicx}
   \graphicspath{{../pdf/}{../jpeg/}}
   \DeclareGraphicsExtensions{.pdf,.jpeg,.png}
\else
   \graphicspath{{../eps/}}
   \DeclareGraphicsExtensions{.eps}
\fi
\begin{document}
\title{SySeVR: A Framework for Using Deep Learning to Detect Software Vulnerabilities}

\author{Zhen~Li, %~\IEEEmembership{Member,~IEEE,}
        Deqing~Zou, %~\IEEEmembership{Fellow,~OSA,}
        Shouhuai~Xu,
        Hai~Jin, ~\IEEEmembership{Fellow,~IEEE,}
        Yawei~Zhu,
        and Zhaoxuan~Chen
%        Sujuan Wang,
%        and~Jialai Wang
        %and~Jane~Doe,~\IEEEmembership{Life~Fellow,~IEEE}% <-this % stops a space
\thanks{Corresponding author: Deqing Zou.}
% <-this % stops a space
\IEEEcompsocitemizethanks{\IEEEcompsocthanksitem Z. Li is with the National Engineering Research Center for Big Data Technology and System, Services Computing Technology and System Lab, Cluster and Grid Computing Lab, Big Data Security Engineering Research Center, School of Cyber Science and Engineering, Huazhong University of Science and Technology, Wuhan 430074, China, and also with School of Cyber Security and Computer, Hebei University, Baoding, 071002, China. 
%\protect\\
E-mail: lizhenhbu@gmail.com
\IEEEcompsocthanksitem D. Zou is with the National Engineering Research Center for Big Data Technology and System, Services Computing Technology and System Lab, Cluster and Grid Computing Lab, Big Data Security Engineering Research Center, School of Cyber Science and Engineering, Huazhong University of Science and Technology, Wuhan 430074, China. 
E-mail: deqingzou@hust.edu.cn
\IEEEcompsocthanksitem S. Xu is with the Department of Computer Science, University of Colorado Colorado Springs, Colorado, USA 80918. This work was done when he was at
University of Texas at San Antonio. 
E-mail: sxu@uccs.edu.
\IEEEcompsocthanksitem H. Jin, Y. Zhu, and Z. Chen are with the National Engineering Research Center for Big Data Technology and System, Services Computing Technology and System Lab, Cluster and Grid Computing Lab, Big Data Security Engineering Research Center, School of Computer Science and Technology, Huazhong University of Science and Technology, Wuhan 430074, China. 
E-mail: \{hjin, yokisir, zhaoxaunchen\}@hust.edu.cn}
% <-this % stops an unwanted space
}


\markboth{IEEE TRANSACTIONS ON DEPENDABLE AND SECURE COMPUTING}
{LI \MakeLowercase{\textit{et al.}}: S\MakeLowercase{y}S\MakeLowercase{e}VR: A Framework for Using Deep Learning to Detect Software Vulnerabilities}



\IEEEtitleabstractindextext{%
\begin{abstract}
The detection of software vulnerabilities (or vulnerabilities for short) is an important problem that has yet to be tackled, as manifested by the many vulnerabilities reported on a daily basis. This calls for machine learning methods for vulnerability detection. Deep learning is attractive for this purpose because it alleviates the requirement to manually define features. Despite the tremendous success of deep learning in other application domains, its applicability to vulnerability detection is not systematically understood. In order to fill this void, we propose the {\em first} systematic framework for using deep learning to detect vulnerabilities in C/C++ programs with source code. The framework, dubbed {\em \underline{Sy}ntax-based, \underline{Se}mantics-based, and \underline{V}ector \underline{R}epresentations} (SySeVR), focuses on obtaining program representations that can accommodate syntax and semantic information pertinent to vulnerabilities. Our experiments with 4 software products demonstrate the usefulness of the framework: we detect 15 vulnerabilities that are not reported in the National Vulnerability Database. Among these 15 vulnerabilities, 7 are unknown

and have been reported to the vendors, and the other 8 have been ``silently'' patched by the vendors when releasing newer versions of the pertinent software products.
\end{abstract}

\begin{IEEEkeywords}
Vulnerability detection, security, deep learning, program analysis, program representation.
\end{IEEEkeywords}}

\maketitle

\IEEEdisplaynontitleabstractindextext

% \fi
\IEEEpeerreviewmaketitle

\section{Introduction}
\label{sec:Introduction}
\IEEEPARstart{S}{oftware} vulnerabilities (or vulnerabilities for short) are a fundamental reason for the prevalence of cyber attacks. Despite academic and industrial efforts at improving software quality, vulnerabilities remain a big problem.
This can be justified by the fact that each year, many vulnerabilities are reported in the {\em Common Vulnerabilities and Exposures} (CVE) \cite{CVE}.

Given that vulnerabilities are inevitable, it is important to detect them as early as possible.
%This is known as the problem of {\em vulnerability detection}.
Source code-based static analysis is an important approach to detecting vulnerabilities, including
{\em code similarity-based} methods \cite{kim2017vuddy,li2016vulpecker} and {\em pattern-based} methods \cite{FlawFinder,RATS,Checkmarx,grieco2016toward,neuhaus2007predicting,yamaguchi2013chucky,yamaguchi2012generalized}. Code similarity-based methods can detect vulnerabilities that are incurred by code cloning, but have high false-negatives when vulnerabilities are not caused by code cloning \cite{vuldeepecker}. Pattern-based methods may require human experts to define vulnerability features for representing vulnerabilities, which makes them error-prone and laborious.
%Moreover, it is also difficult to achieve both low false-positive rate and low false-negative rate.
Therefore, an ideal method should be able to effectively detect vulnerabilities caused by a wide range of reasons while imposing as little reliance on human experts as possible.

Deep learning --- including {\em Recurrent Neural Networks} (RNNs) \cite{DBLP:conf/icse/0004CC17, white2016deep, shin2015recognizing, DBLP:conf/ccs/LinZLPX17}, %DBLP:conf/esorics/AlsulamiDHMG17},
{\em Convolutional Neural Networks} (CNNs) \cite{DBLP:conf/qrs/LiHZL17, DBLP:journals/chinaf/GengZC18, DBLP:journals/chinaf/WangSLDZ17}, and {\em Deep Belief Networks} (DBNs) \cite{wang2016automatically, DBLP:conf/qrs/YangLXZS15} ---
has been successful in image and natural language processing.
While it is tempting to use deep learning to detect vulnerabilities, we observe that there is a ``domain gap'':
deep learning is born to cope with data with natural {\em vector representations} (e.g., pixels of images);
in contrast, software programs do not have such vector representations.
% that are suitable for deep learning-based vulnerability detection.
%that are the required input for deep neural networks.
Recently, we proposed the {\em first} deep learning-based vulnerability detection system, dubbed VulDeePecker \cite{vuldeepecker}, to detect vulnerabilities at the slice level (i.e., multiple lines of code that are semantically related to each other).
While demonstrating the feasibility of using deep learning to detect vulnerabilities, VulDeePecker has four weaknesses:
(i) it considers only the vulnerabilities that are related to library/API function calls;
(ii) it leverages only the semantic information induced by {\em data dependency};
(iii) it considers only a particular RNN known as {\em Bidirectional Long Short-Term Memory} (BLSTM); and
(iv) it makes no effort to explain the cause of false-positives and false-negatives.

\ignore{

The limitations mentioned above motivates the present study, which aims to answer the following questions:
\begin{itemize}
\item Can deep neural networks be used to detect all kinds of vulnerabilities?
\item Which deep neural network is more effective in detecting vulnerabilities?
\item Can control-flow information improve the detection capability of a deep neural network? If so, to what extent?
\end{itemize}

}

\smallskip
\noindent{\bf Our contributions.}
In this paper, we propose the {\em first} systematic framework for using deep learning to detect vulnerabilities in C/C++ programs with source code. The framework is centered at answering the following question: {\em How can we represent programs as vectors that accommodate the syntax and semantic information that is suitable for vulnerability detection?}
In order to answer this question, we introduce the notions of {\em \underline{Sy}ntax-based \underline{V}ulnerability \underline{C}andidates} (SyVCs) and {\em \underline{Se}mantics-based \underline{V}ulnerability \underline{C}andidates} (SeVCs). Intuitively, SyVCs reflect vulnerability syntax characteristics, and SeVCs extend SyVCs to accommodate the semantic information induced by data dependency and control dependency. Moreover, we design algorithms to extract SyVCs and SeVCs automatically. This explains why we call the framework \underline{Sy}ntax-based, \underline{Se}mantics-based, and \underline{V}ector  \underline{R}epresentations, or SySeVR for short.
%, because it considers syntax-based, semantics-based, and vector representations.
As we will see, SySeVR overcomes the aforementioned weaknesses (i)-(iv) of VulDeePecker \cite{vuldeepecker}.

In order to evaluate the effectiveness of SySeVR, we present a dataset of 126 types of vulnerabilities, which are collected
from the {\em National Vulnerability Database} (NVD) \cite{NVD} and the {\em Software Assurance Reference Dataset} (SARD) \cite{SARD}.
This dataset should be of independent value and is made publicly available at \url{https://github.com/SySeVR/SySeVR}.
It is worth mentioning that the dataset we published earlier in association to VulDeePecker \cite{vuldeepecker} is not sufficient for the purpose of the present paper, simply because the dataset associated to \cite{vuldeepecker} contains only 2 types of vulnerabilities.

Equipped with the new dataset, we show that SySeVR achieves the following.

\begin{itemize}
\item SySeVR enables multiple kinds of neural networks to detect various kinds of vulnerabilities.
In the SySeVR framework, Bidirectional RNNs, especially {\em Bidirectional Gated Recurrent Unit} (BGRU),
%(especially BGRU)
%more effective than CNNs, and makes CNNs
are more effective than unidirectional RNNs and CNNs, which are more effective than DBNs and shallow learning models.
Moreover, SySeVR makes deep neural networks (especially BGRU) much more effective than the state-of-the-art vulnerability detection methods.
\item The effectiveness of BGRU is substantially affected by the training data.
If some syntax elements (e.g., tokens) often appear in vulnerable (vs. not vulnerable) pieces of code, then these syntax elements may cause high false-positive rates (correspondingly, false-negative rates).
This means that we can explain the cause of false-positives and false-negatives to some extent.
\item Accommodating more semantic information (i.e., control dependency and data dependency) can improve the effectiveness of SySeVR-enabled vulnerability detectors.
For example, semantic information induced by data dependency and control dependency can reduce the false-negative rate by 30.4\% on average.
\item By applying SySeVR-enabled BGRU to 4 software products (Libav, Seamonkey, Thunderbird, and Xen), we detect 15 vulnerabilities that have not been reported in NVD \cite{NVD}. Among these 15 vulnerabilities, 7 are unknown to exist in these software products;
% despite that similar vulnerabilities are known to exist in other software);
for ethical reasons, we do {\em not} release the precise locations of these vulnerabilities, but we have reported them to the respective vendors. The other 8 vulnerabilities have been ``silently'' patched by the vendors when releasing newer versions of the pertinent software products.
\end{itemize}

%\smallskip
\noindent{\bf Paper outline.}
Section \ref{sec:Design} presents the SySeVR framework.
Section \ref{sec:Experimental_results} describes experiments and results.
Section \ref{sec:Limitations} discusses limitations of the present study.
Section \ref{sec:Related_work} reviews related prior work.
Section \ref{sec:Conclusion} concludes the paper.

\section{The SySeVR Framework}
\label{sec:Design}
\subsection{Basic Idea and Framework Overview}

\begin{figure*}[!htbp]
\vspace{-0.2cm}
\centering
\includegraphics[width=0.75\textwidth]{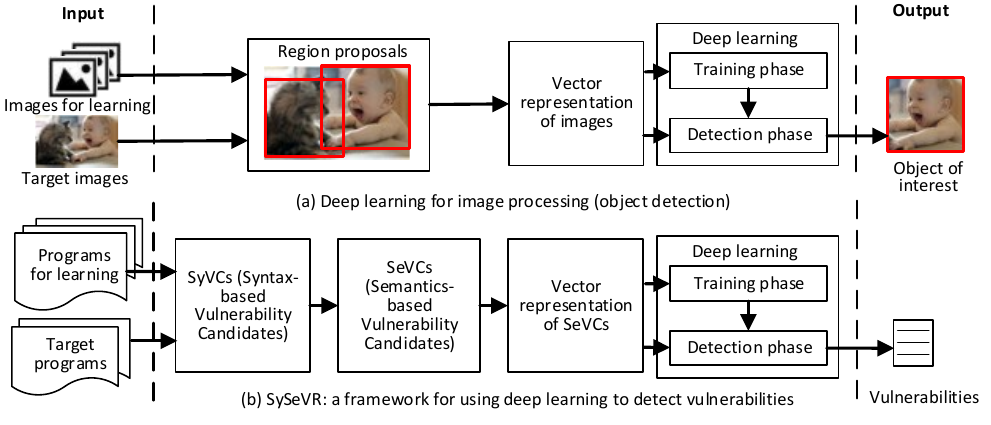}
\vspace{-0.2cm}
\caption{
(a) The notion of {\em region proposal} in image processing. (b) The SySeVR framework
%for using deep learning to detect vulnerabilities, which is
is inspired by the notion of region proposal and is centered on obtaining SyVC, SeVC, and vector representations of programs.}
\label{Fig_Overview_of_our_approach}
\end{figure*}

\subsubsection{Basic Idea}
Deep learning is successful in image processing and other applications.
%In order to clearly see the gap between these domains (i.e., the ``domain gap''),
In particular, the notion of {\em region proposal} \cite{ren2015faster,shrivastava2016training} in image processing inspires us to adapt it to the context of vulnerability detection. 
However, the problem vulnerability detection is very different from the problem image processing because the latter has natural structural representations.
To see the difference, let us consider an example of using deep learning to detect humans in images. On one hand, as illustrated in Fig. \ref{Fig_Overview_of_our_approach}(a), detecting humans in an image can be achieved by using the notion of {\em region proposal} and leveraging the structural representation of images (e.g., texture, edge, and color). Multiple region proposals can be extracted from an image, and each region proposal can be treated as a ``unit'' for training a neural network to detect objects (i.e., humans in this example).

On the other hand, when using deep learning to detect vulnerabilities, we need to represent programs in a way that can adequately accommodate the syntax and semantic information related to vulnerabilities.
At a first glance, one may suggest treating each {\em function} in a program as a region proposal in image processing. However, this is too coarse-grained because vulnerability detectors not only need to tell whether a function is vulnerable or not, but also need to pin down locations of vulnerabilities. That is, we need fine-grained representations of programs for vulnerability detection. One may also suggest treating each {\em line of code} or {\em statement} (i.e., these two terms will be used interchangeably) as a unit for vulnerability detection. However, this treatment has two drawbacks: (i) most statements in a program do not contain any vulnerability, meaning that few samples are vulnerable; and (ii) multiple statements that are semantically related to each other are not considered as a whole.

The preceding discussion suggests us to divide a program into smaller pieces of code (i.e., a number of statements),
which correspond to ``region proposals'' and exhibit the {\em syntax} and {\em semantics} characteristics of vulnerabilities.

\begin{figure*}[!htbp]
\vspace{-0.2cm}
\centering
\includegraphics[width=0.9\textwidth]{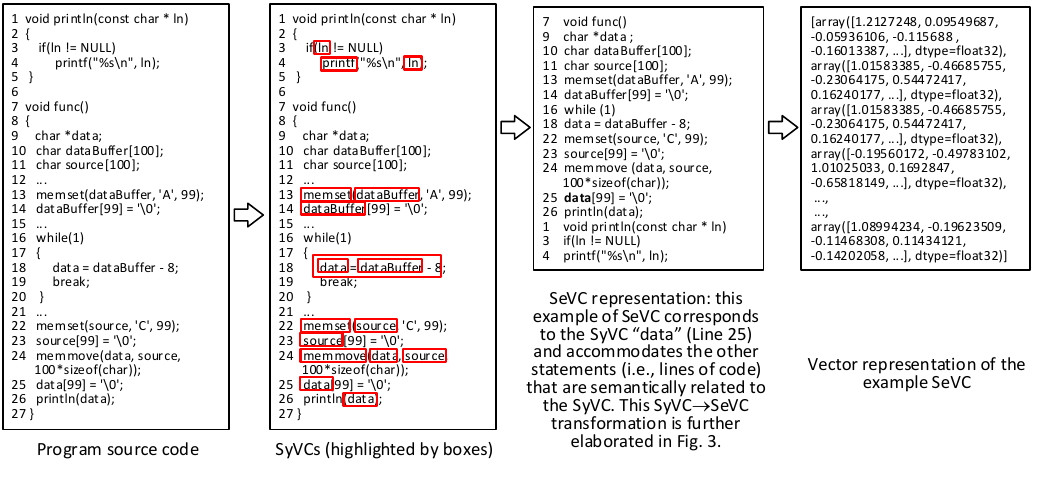}
\vspace{-0.2cm}
\caption{An example illustrating SyVC, SeVC, and vector representations of program, where SyVCs are highlighted by boxes and one SyVC may be part of another SyVC. The SyVC$\to$SeVC transformation is elaborated in Fig. \ref{Fig_generating_code_gadgets}.}
\vspace{-0.2cm}
\label{Fig_three_representation}
\end{figure*}

\smallskip

\subsubsection{Framework Overview}

We observe that vulnerabilities exhibit some {\em syntax characteristics}, such as function call or pointer usage.
% (see in Section \ref{sec:Experimental_results}).
Therefore, we propose using syntax characteristics to identify SyVCs, which serve as a {\em starting point} for vulnerability detection (i.e., SyVCs are {\em not} sufficient for training deep learning models because they accommodate {\em no} semantic information of vulnerabilities).
Fig. \ref{Fig_Overview_of_our_approach}(b) highlights the SySeVR framework inspired by the notion of region proposal. Essentially, the framework seeks SyVC, SeVC, and vector representations of programs that are suitable for vulnerability detection.

In order to help understand SySeVR, we use the running example described in Fig. \ref{Fig_three_representation} to highlight how SySeVR extracts SyVC, SeVC, and vector representations of programs. At a high level, a SyVC, which is highlighted by a box in Fig. \ref{Fig_three_representation}, is a code element that 
%may or may not be vulnerable according to 
matches the syntax characteristics of  some vulnerability. A SeVC extends a SyVC to include statements (i.e., lines of code) that are semantically related to the SyVC,
where semantic information is induced by
control dependency and/or data dependency; this ``SyVC to SeVC'' (or SyVC$\to$SeVC) transformation is fairly involved and will be 
%therefore
elaborated later (in Fig. \ref{Fig_generating_code_gadgets}). Finally, each SeVC is encoded into a vector for input to a deep neural network.

\subsection{Extracting SyVCs}

\subsubsection{Extracting Vulnerability Syntax Characteristics}

We propose using vulnerability syntax characteristics to identify pieces of code as {\em initial candidates} for vulnerability detection.
For example, vulnerabilities associated to pointer usage would exhibit that the declaration of an identifier contains character `$\ast$'.
In Fig. \ref{Fig_three_representation}, the identifier ``data'' in Line 18 of the program source code is a pointer usage, and the declaration of ``data'' in Line 9 contains a character `$\ast$'.

Given that there are many vulnerabilities, we anticipate it to be extremely time-consuming to define and extract their syntax characteristics because this requires to extract the vulnerable lines of code from the vulnerable programs. While this itself is an important research problem, in Section \ref{subsubsection:extracting-SyVCs} we will propose a specific method for extracting vulnerability syntax characteristics; we note that this method is far from perfect because it only covers 93.6\% of the vulnerable programs we collected, but is sufficient for demonstrating the usefulness of SySeVR. 
In our method, we use the attributes of nodes on the {\em Abstract Syntax Tree} (AST) of a program to describe vulnerability syntax characteristics.

Regardless of the specific descriptions of vulnerability syntax characteristics, we can use $H=\{h_k\}_{1\leq k \leq \beta}$ to denote a set of vulnerability syntax characteristics, where $h_k$ represents a vulnerability syntax characteristic and $\beta$ is the number of vulnerability syntax characteristics.
Given $H$,
%$H=\{h_k\}_{1\leq k \leq \beta}$ 
we need to determine whether a piece of code matches a syntax characteristic $h_k$ or not. Since these matching operations are specific to the representation of vulnerability syntax characteristics, we defer their description to our case study with a specific representation of vulnerability syntax characteristics.

{\color{black}\subsubsection{Defining and Extracting SyVCs}}
%In order to precisely define what SyVCs are,
We start with the definition of programs, functions, statements and tokens that will be used throughout of the present paper.

\begin{definition}[program, function, statement, token]
\label{definition:program}
%\emph{(program, function, statement, token)}
A program $P$ is a set of functions $f_1, \ldots,f_\eta$, denoted by $P=\{f_1, \ldots,f_\eta\}$.
A function $f_i$, where $1\leq i \leq \eta$, is an ordered set of statements $s_{i,1}, \ldots,s_{i,m_i}$, denoted by $f_i=\{s_{i,1}, \ldots,s_{i,m_i}\}$.
%where $s_{i,j}$, $1\leq j \leq m_i$, is a statement. %(i.e., line of code).
A statement $s_{i,j}$, where $1\leq i \leq \eta$ and $1\leq j \leq m_i$, is an ordered set of tokens $t_{i,j,1},\ldots,t_{i,j,w_{i,j}}$, denoted by $s_{i,j}=\{t_{i,j,1},\ldots,t_{i,j,w_{i,j}}\}$. Note that tokens can be identifiers, operators, constants, and keywords, and can be extracted by lexical analysis.
\end{definition}

Given a function $f_i$, there are standard routines for generating its AST
%Abstract Syntax Tree (AST) 
\cite{DBLP:journals/cl/Noonan85}. %\footnote{add reference}
%which is denoted by $T_i$.
%On $T_i$,
The root of the AST corresponds to function $f_i$, a leaf of the AST corresponds to a token $t_{i,j,g}$ ($1\leq g \leq w_{i,j}$),
%and an internal node corresponds to a statement $s_{i,j}$.
and an internal node of the AST corresponds to a statement $s_{i,j}$ or multiple consecutive tokens of $s_{i,j}$.
Intuitively, a SyVC corresponds to a leaf node of an AST, meaning that it is a token, or corresponds to an internal node of an AST, meaning that it is a statement or consists of multiple consecutive tokens.
% or consists of multiple consecutive tokens ().
Formally,

\begin{definition}[SyVC]
Consider a program $P=\{f_1, \ldots$, $f_\eta\}$, where $f_i=\{s_{i,1}, \ldots,s_{i,m_i}\}$ with $s_{i,j}=\{t_{i,j,1},\ldots$, $t_{i,j,w_{i,j}}\}$.
A {\em code element} $e_{i,j,z}$ is composed of one or multiple consecutive tokens of $s_{i,j}$, namely $e_{i,j,z}=(t_{i,j,u},\ldots,t_{i,j,v})$ where $1\leq u\leq v\leq w_{i,j}$.
Given a set of vulnerability syntax characteristics $H=\{h_k\}_{1\leq k \leq \beta}$, where $h_k$ represents a vulnerability syntax characteristic
and $\beta$ is the number of vulnerability syntax characteristics as mentioned above, a code element $e_{i,j,z}$ that {\em matches} a vulnerability syntax characteristic $h_k$ is called a SyVC, where the ``matching'' operation, as discussed above, is related to the specific representation of vulnerability syntax characteristics.
%For example, $h_k$ for vulnerabilities related to pointer usage is the usage of an identifier whose declaration contains a character `$\ast$'.
\end{definition}

Algorithm \ref{alg_obtaining_SyVC} gives a high-level description on the extraction of SyVCs from a given program $P=\{f_1, \ldots,f_\eta\}$ and a set $H=\{h_k\}_{1\leq k\leq \beta}$ of vulnerability syntax characteristics.
Specifically, Algorithm \ref{alg_obtaining_SyVC} uses a standard routine to generate an AST $T_i$ for each function $f_i$.
Then, Algorithm \ref{alg_obtaining_SyVC} traverses $T_i$ to identify SyVCs, namely the code elements that ``match'' some $h_k$,
where the ``matching'' operation is related to the representation of vulnerability syntax characteristics and therefore will be elaborated when coping with specific vulnerability syntax characteristics (see Section \ref{subsubsection:extracting-SyVCs}).

\begin{algorithm}[h]
%\begin{algorithm}[htbp!]
%\algsetup{linenosize=\scriptsize}
\footnotesize
\caption{Extracting SyVCs from a program}
\label{alg_obtaining_SyVC}
\begin{basedescript}{\desclabelstyle{\pushlabel}\desclabelwidth{6em}}
\item[Input:]
A program $P=\{f_1, \ldots,f_\eta\}$; a set $H=\{h_k\}_{1 \leq k\leq \beta}$ of vulnerability syntax characteristics
\item[Output:]
A set $Y$ of SyVCs
\end{basedescript}
\begin{algorithmic}[1]
\STATE $Y \leftarrow \emptyset$;
\FOR{each function $f_i \in P$}
    \STATE Generate an abstract syntax tree $T_i$ for $f_i$;
    \FOR{each code element $e_{i,j,z}$ in $T_i$}
    %\FOR{each node $\alpha_{i,j,z}$ in $T_i$}
        \FOR{each $h_k \in H$}
            %\IF{code element $e_{i,j,z}$ corresponding to $\alpha_{i,j,z}$ satisfies the syntax characteristics of $h_k$}
            \IF{$e_{i,j,z}$ matches $h_k$} % (see Section \ref{sec:Experimental_results} for details)}
            %and $e_{i,j,z}$ is the code element corresponding to $\alpha_{i,j,z}$~~
                \STATE $Y \leftarrow Y \cup$ \{$e_{i,j,z}$\};
            \ENDIF
        \ENDFOR
    \ENDFOR
\ENDFOR
\RETURN $Y$; \COMMENT{the set of SyVCs}
\end{algorithmic}
\end{algorithm}
%\vspace{-0.6cm}

In order to help understand the idea, we now consider an example.
In the second column of Fig. \ref{Fig_three_representation}, we use boxes to highlight all of the SyVCs that are extracted from the program source code using the vulnerability syntax characteristics that will be described in Section \ref{subsubsection:extracting-SyVCs}. We will elaborate how these SyVCs are extracted. It is worth mentioning that one SyVC may be part of another SyVC. For example, there are three SyVCs that are extracted from Line 18 because they are extracted with respect to different vulnerability syntax characteristics.
%This manifests that different SyVCs may correspond to the same type of vulnerabilities.

{\color{black}\subsection{Transforming SyVCs to SeVCs}}

{\color{black}\subsubsection{Basic Definitions}}
In order to detect vulnerabilities, we propose transforming SyVCs to SeVCs (i.e., SyVC$\to$SeVC) to accommodate the statements that are semantically related to the SyVCs in question.
For this purpose, we propose leveraging the {\em program slicing} technique to identify the statements that are semantically related to SyVCs.
In order to use the program slicing technique, we need to use {\em Program Dependency Graph} (PDG).
This requires us to use {\em data dependency} and {\em control dependency}, which are defined over {\em Control Flow Graph} (CFG).
These concepts are reviewed below.

\begin{definition}[CFG \cite{DBLP:journals/toplas/FerranteOW87}]
%\emph{(CFG \cite{DBLP:journals/jpl/Tip95})}
For a program $P=\{f_1, \ldots$, $f_\eta\}$,
the CFG of function $f_i$ is a graph $G_i=(V_i, E_i)$, where $V_i=\{n_{i,1}, \ldots, n_{i,c_i}\}$ is a set of nodes with each node representing a statement or control predicate,
and $E_i=\{\epsilon_{i,1}, \ldots$, $\epsilon_{i,d_i}\}$  is a set of direct edges with each edge representing the possible flow of control between a pair of nodes.
\end{definition}

\begin{definition}[data dependency \cite{DBLP:journals/toplas/FerranteOW87}]
%\emph{(data dependency \cite{DBLP:journals/jpl/Tip95})}
Consider a program $P=\{f_1, \ldots,f_\eta\}$, the CFG $G_i=(V_i,E_i)$ of function $f_i$, and two nodes $n_{i,j}$ and $n_{i,\ell}$ in $G_i$ where $1 \leq j, \ell \leq c_i$ and $j \neq \ell$.
If there is a path from $n_{i,\ell}$ to $n_{i,j}$ in $G_i$ and a value computed at node $n_{i,\ell}$ is used at node $n_{i,j}$, then $n_{i,j}$ is {\em data-dependent} on $n_{i,\ell}$.
\end{definition}

\begin{definition}[control dependency \cite{DBLP:journals/toplas/FerranteOW87}]
%\emph{(control dependency \cite{DBLP:journals/jpl/Tip95})}
Consider a program $P=\{f_1, \ldots,f_\eta\}$, the CFG $G_i=(V_i,E_i)$ of function $f_i$, and two nodes $n_{i,j}$ and $n_{i,\ell}$ in $G_i$ where $1 \leq j, \ell \leq c_i$ and $j \neq \ell$.
It is said that $n_{i,j}$ {\em post-dominates} $n_{i,\ell}$ if all paths from $n_{i,\ell}$ to the end of the program traverse through $n_{i,j}$.
If there exists a path starting at $n_{i,\ell}$ and ending at $n_{i,j}$ such that (i) $n_{i,j}$ {\em post-dominates} every node on the path excluding $n_{i,\ell}$ and $n_{i,j}$, and (ii) $n_{i,j}$ does not post-dominate $n_{i,\ell}$, then $n_{i,j}$  is {\em control-dependent} on $n_{i,\ell}$.
\end{definition}

Based on data dependency and control dependency, PDG can be defined as follows.

\begin{definition}[PDG \cite{DBLP:journals/toplas/FerranteOW87}]
%\emph{(PDG \cite{DBLP:journals/jpl/Tip95})}
For a program $P=\{f_1, \ldots$, $f_\eta\}$,
the PDG of function $f_i$ is denoted by $G_i'=(V_i,E_i')$, where $V_i$ is the same as in CFG $G_i$,
%'=\{n_{i,1}', \ldots, n_{i,c_i'}'\}$ is a set of nodes with each node representing a statement or control predicate,
and $E_i'=\{\epsilon_{i,1}', \ldots$, $\epsilon_{i,d_i'}'\}$  is a set of direct edges with each edge representing a data or control dependency between a pair of nodes.
\end{definition}

\subsubsection{Defining Program Slices}
Given PDGs, we can extract {\em program slices}
%{\em interprocedural program slices}
from SyVCs.
%which may go beyond the boundaries of individual functions.
We consider both forward and backward slices because (i) a SyVC may affect some subsequential statements, which may therefore contain a vulnerability; and (ii) the statements affecting a SyVC may render the SyVC vulnerable. Formally,

\begin{definition}[forward, backward, and program slices \cite{DBLP:journals/jpl/Tip95} of a SyVC]
Consider a program $P=\{f_1, \ldots,f_\eta\}$, the PDG $G_i'=(V_i,E_i')$ for each function $f_i$ ($1\leq i \leq \eta$), and a SyVC, $e_{i,j,z}$, of statement $s_{i,j}$ in $G_i'$.
\begin{itemize}
\item The {\em forward slice} of SyVC $e_{i,j,z}$ in $f_i$, denoted by $\fs_{i,j,z}$, is defined as an ordered set of nodes $\{n_{i,x_1}$, $\ldots$, $n_{i,x_{\mu_i}}\}\subseteq V_i$, %in $G_i'$,
    where $n_{i, x_p}$, $1\leq x_1 \leq x_p \leq x_{\mu_i} \leq c_i$, is reachable from $e_{i,j,z}$ in $G_i'$.
    That is, the nodes in $\fs_{i,j}$ are from all paths in $G_i'$ starting at $e_{i,j,z}$.

\item The {\em interprocedural forward slice} of SyVC $e_{i,j,z}$ in program $P$,
denoted by ${\sf fs}'_{i,j,z}$, is defined as an ordered set of nodes,
where (i) a node belongs to one or multiple PDGs and (ii) each node is reachable starting from $e_{i,j,z}$ via a sequence of function calls.
That is, ${\sf fs}'_{i,j,z}$ is a forward slice with or without crossing function boundaries
%going beyond function boundaries
(via function calls).

\item The {\em backward slice} of SyVC $e_{i,j,z}$ in $f_i$, denoted by $\bs_{i,j,z}$, is defined as an ordered set of nodes $\{n_{i,y_1}, \ldots,$ $n_{i,y_{\nu_i}}\}\subseteq V_i$, where $n_{i, y_p}$, $1\leq y_1 \leq y_p \leq y_{\nu_i} \leq c_i$, from which $e_{i,j,z}$ is reachable in $G_i'$.
That is, the nodes in $\bs_{i,j,z}$ are from all
%possible
paths in $G_i'$ ending at $e_{i,j,z}$.

\item The {\em interprocedural backward slice} of SyVC $e_{i,j,z}$ in program $P$, denoted by ${\sf bs}'_{i,j,z}$,
    is defined as an ordered set of nodes,
% in $G_i'$,\footnote{or in the union of the $G_i'$'s?}
where (i) a node belongs to one or multiple PDGs and (ii) each node can reach $e_{i,j,z}$ via a sequence of function calls.
%(e.g., $f_i$ calls $f_{a_p}$ where $i \neq a_p$).
That is, ${\sf bs}'_{i,j,z}$ is a backward slice with or without crossing function boundaries
%going beyond function boundaries
(via function calls).

\item Given an interprocedural forward slice $\fs_{i,j,z}'$ and an interprocedural backward slice $\bs_{i,j,z}'$, the (interprocedural) {\em  program slice}
of SyVC $e_{i,j,z}$, denoted by ${\sf ps}_{i,j,z}$, is defined as an ordered set of nodes (belonging to the PDGs of functions in $P$) by merging $\fs_{i,j,z}'$ and $\bs_{i,j,z}'$ at the SyVC $e_{i,j,z}$.
That is, ${\sf ps}_{i,j,z}$ is an ordered set obtained
%the ordered set union of $\fs_{i,j,z}'$ and $\bs_{i,j,z}'$, 
by connecting forward slice $\fs_{i,j,z}'$ and backward slice $\bs_{i,j,z}'$ in an order-preserving fashion while omitting the adjacent repeating nodes (i.e., using one node to replace the multiple adjacent appearances of the same node). 
%denoted by $\fs_{i,j,z}' \cup \bs_{i,j,z}'$.
\end{itemize}
\end{definition}

In Fig. \ref{Fig_generating_code_gadgets}, the third column shows the interprocedural forward slice, the interprocedural backward slice, and the program slice of SyVC ``data'' (Line 25 in the program source code). The interprocedural forward slice of SyVC ``data'' crosses functions $func$ and $println$. 
The interprocedural backward slice of SyVC ``data'' is the same as the backward slice of SyVC ``data'' in function $func$, 
%The interprocedural backward slice of SyVC ``data'' is a backward one 
because there is no other function that calls function $func$.
The program slice of SyVC ``data'' is obtained by connecting the interprocedural forward slice and the interprocedural backward slice while omitting one (of the two) adjacent appearance of the node corresponding to SyVC ``data'' (Line 25 in the program source code).

\subsubsection{Defining SeVCs}
Having extracted program slices of SyVCs, we can now define SeVCs.

\begin{definition}[SeVC]
%\emph{(SeVC)}
Given a program $P=\{f_1, \ldots,f_\eta\}$ and a SyVC $e_{i,j,z}$ in statement $s_{i,j}$ of function $f_i$,
the SeVC corresponding to SyVC $e_{i,j,z}$, denoted by  $\delta_{i,j,z}$, is defined as an ordered subset of statements in $P$, denoted by $\delta_{i,j,z}=\{s_{a_1,b_1}, \ldots, s_{a_{v_{i,j,z}},b_{v_{i,j,z}}}\}$,
where a data dependency or control dependency exists between statement $s_{a_p,b_q}$ ($1 \leq p, q \leq v_{i,j,z}$)
and SyVC $e_{i,j,z}$.
% that are in a same function or in two different functions.
In other words, a SeVC $\delta_{i,j,z}$ is an ordered set of statements that correspond to the nodes of (interprocedural) program slice ${\sf ps}_{i,j,z}$.
\end{definition}

\begin{algorithm}[!htb]
%\algsetup{linenosize=\scriptsize}
\footnotesize
\caption{Transforming SyVCs to SeVCs}
\label{alg_obtaining_ SeVC}
\begin{basedescript}{\desclabelstyle{\pushlabel}\desclabelwidth{6em}}
\item[Input:]
A program $P=\{f_1,\ldots,f_\eta\}$;\\
a set $Y$ of SyVCs generated by Algorithm \ref{alg_obtaining_SyVC}
\item[Output:]
The set of SeVCs
\end{basedescript}
\begin{algorithmic}[1]
\STATE $C \leftarrow \emptyset$;
\FOR{each $f_i \in P$} \label{algorithm2:begin-step1}
    \STATE Generate a PDG $G_i'=(V_i,E_i')$ for $f_i$;
\ENDFOR  \label{algorithm2:end-step1}
\FOR{each $e_{i,j,z} \in Y$ in $G_i'$}
    \STATE{Generate forward slice $\fs_{i,j,z}$ \& backward slice $\bs_{i,j,z}$ of $e_{i,j,z}$; \label{algorithm2:begin-step2}}
          \STATE{Generate interprocedural forward slice ${\sf fs}_{i,j,z}'$ by interconnecting $\fs_{i,j,z}$ and the forward slices from the functions called by $f_i$;}
          \STATE{Generate interprocedural backward slice ${\sf bs}_{i,j,z}'$ by interconnecting $\bs_{i,j,z}$ and the backward slices from both the functions called by $f_i$ and the functions calling $f_i$;}
          \STATE{Generate program slice ${\sf ps}_{i,j,z}$ by connecting $\fs_{i,j,z}'$ and $\bs_{i,j,z}'$ at $e_{i,j,z}$;}\label{algorithm2:end-step2}
    \FOR{each statement $s_{i,j} \in f_i$ appearing in ${\sf ps}_{i,j,z}$ as a node}  \label{algorithm2:begin-step3}
        \STATE $\delta_{i,j,z} \leftarrow \delta_{i,j,z} \cup \{s_{i,j}\}$, according to the order of the appearance of $s_{i,j}$ in $f_i$;
    \ENDFOR
  \FOR{two statements $s_{i,j} \in f_i$ and $s_{a_p,b_q} \in f_{a_p}$ ($i \neq a_p$) appearing in ${\sf ps}_{i,j,z}$ as nodes}
       \IF{$f_i$ calls $f_{a_p}$}
          \STATE  $\delta_{i,j,z} \leftarrow \delta_{i,j,z} \cup \{s_{i,j}, s_{a_p,b_q}\}$, where $s_{i,j}<s_{a_p,b_q}$;
       \ELSE
          \STATE  $\delta_{i,j,z} \leftarrow \delta_{i,j,z} \cup \{s_{i,j}, s_{a_p,b_q}\}$, where $s_{i,j}>s_{a_p,b_q}$;
          %according to the order in a slice, or the random order otherwise;
       \ENDIF
  \ENDFOR   \label{algorithm2:end-step3}
\STATE  $C \leftarrow C \cup \{\delta_{i,j,z}\}$;
\ENDFOR
%\STATE Delete duplications in $C$;
\RETURN $C$; \COMMENT{the set of SeVCs}
\end{algorithmic}
\end{algorithm}

\begin{figure*}[!htbp]
%\vspace{-0.2cm}
\centering
\includegraphics[width=.9\textwidth]{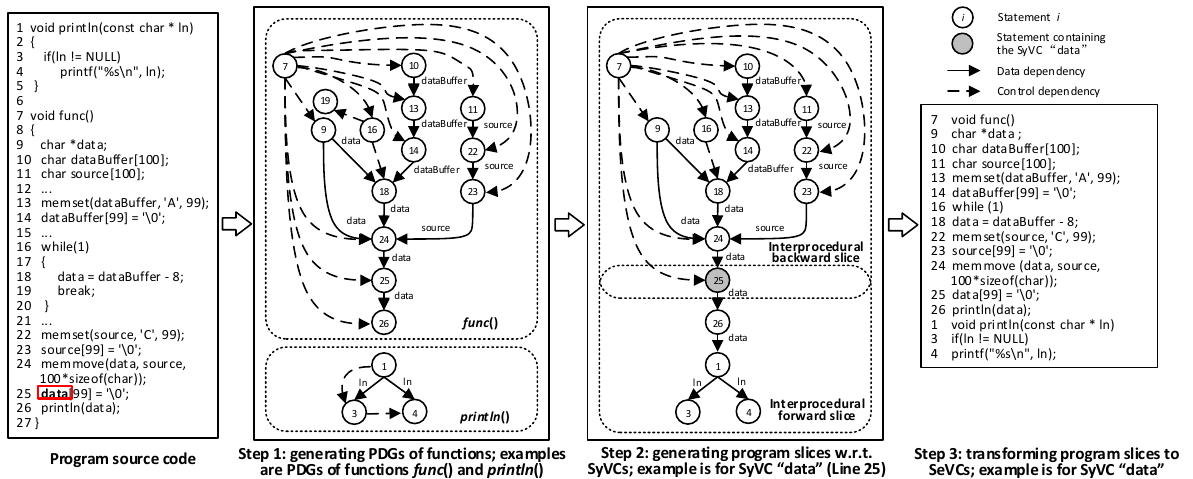}
\vspace{-0.3cm}
\caption{Elaborating the SyVC$\to$SeVC transformation in Algorithm \ref{alg_obtaining_ SeVC} for SyVC ``$data$'', where
solid arrows (i.e., directed edges) represent data dependency, and dashed arrows represent control dependency.
Note that each solid arrow (i.e., data dependency) is annotated with the name of the variable that incurred the data dependency in question.}
\vspace{-0.2cm}
\label{Fig_generating_code_gadgets}
\end{figure*}

{\color{black}\subsubsection{Computing SeVCs}}
Algorithm \ref{alg_obtaining_ SeVC} summarizes the preceding discussion in three steps: generating PDGs; generating program slices of the SyVCs output by Algorithm \ref{alg_obtaining_SyVC}; and transforming program slices to SeVCs.
\ignore{
The computational complexity of the algorithm is $O(\eta m_i^2 w_{i,j} )$, where $\eta$ is the number of functions in program $P$, $m_i$ is the number of statements in function $f_i$, and $w_{i,j}$ is the number of tokens in $s_{i,j}$. }
In what follows we elaborate these steps and use Fig. \ref{Fig_generating_code_gadgets} to illustrate a running example.
Specifically, Fig. \ref{Fig_generating_code_gadgets} elaborates the SyVC$\to$SeVC transformation of SyVC ``$data$'' (related to pointer usage)
%(line 25 of the program in Figure \ref{Fig_three_representation})
while accommodating semantic information induced by data dependency and control dependency.
%Note that in the program, some statements (i.e., lines of code) are omitted because they are not related to the SyVC ``$data$''.

\noindent{\bf Step 1 (Lines \ref{algorithm2:begin-step1}-\ref{algorithm2:end-step1} in Algorithm \ref{alg_obtaining_ SeVC})}. This step generates a PDG for each function. For this purpose, there are standard algorithms (e.g., \cite{DBLP:journals/toplas/FerranteOW87}).
As a running example, the second column of Fig. \ref{Fig_generating_code_gadgets} shows the PDGs respectively corresponding to functions $func$ and $println$, where each number represents the line number of a statement.

\noindent{\bf Step 2 (Lines \ref{algorithm2:begin-step2}-\ref{algorithm2:end-step2} in Algorithm \ref{alg_obtaining_ SeVC})}. This step generates
%forward slice ${\sf fs}_{i,j,z}$ and backward slice $\bs_{i,j,z}$
%the (interprocedural)
the program slice ${\sf ps}_{i,j,z}$ for each SyVC $e_{i,j,z}$.
The interprocedural forward slice $\fs_{i,j,z}'$ is obtained by merging $\fs_{i,j,z}$ and the forward slices from the functions
%of the functions that are
called by $f_i$.
% with call edges that go from a call site to the function entry point.
The interprocedural backward slice $\bs_{i,j,z}'$ is obtained by merging $\bs_{i,j,z}$ and the backward slices
%of the functions that are called by $f_i$ and the backward slices of the functions that call $f_i$.
from both the functions called by $f_i$ and the functions calling $f_i$.
Finally, $\fs_{i,j,z}'$ and $\bs_{i,j,z}'$ are merged into a program slice ${\sf ps}_{i,j,z}$.

As a running example, the third column in Fig. \ref{Fig_generating_code_gadgets} shows the program slice of SyVC ``$data$'', where the backward slice corresponds to function $func$ and the forward slice corresponds to functions $func$ and $println$.
It is worth mentioning that for obtaining the forward slice of a SyVC, we leverage only data dependency for two reasons:
(i) statements affected by a SyVC via control dependency would not be vulnerable in most cases
and (ii) utilizing statements that have a control dependency on a SyVC would involve many statements that have little to do with vulnerabilities.
Consider for example a pointer variable SyVC in the condition expression of a ``while'' loop.
If the pointer variable is not referred to in the body of the ``while'' loop, the statements in the body of the ``while'' loop are
affected by the SyVC only via control dependency, meaning that the SyVC would not cause any vulnerability in the body of the ``while'' loop. If the forward slice of the pointer variable related SyVC mentioned above involves a control dependency, all of the statements in the body of the ``while'' loop, which are control-dependent on the SyVC, would be contained in the SeVC despite that they have little to do with vulnerabilities.
On the other hand, for obtaining the backward slice of a SyVC, we
%have to
leverage both data dependency and control dependency.

\noindent{\bf Step 3 (Lines \ref{algorithm2:begin-step3}-\ref{algorithm2:end-step3} in Algorithm \ref{alg_obtaining_ SeVC})}.
This step transforms program slices to SeVCs as follows.
First, the algorithm transforms the statements belonging to function $f_i$ and appearing in ${\sf ps}_{i,j,z}$ as nodes to a SeVC,
while preserving the order of these statements in $f_i$. As a running example shown in Fig. \ref{Fig_generating_code_gadgets}, 13 statements belong to function $func$, and 3 statements belong to function $println$. According to the order of these statements in the two functions, we obtain two ordered sets of statements: Lines \{7, 9, 10, 11, 12, 14, 16, 18, 22, 23, 24, 25, 26\} and Lines \{1, 3, 4\}.

Second, the algorithm transforms the statements belonging to different functions to a SeVC.
For statements $s_{i,j} \in f_i$  and $s_{a_p,b_q} \in f_{a_p}$ ($i \neq a_p$) appearing in ${\sf ps}_{i,j,z}$ as nodes,
%corresponding to the nodes in ${\sf ps}_{i,j,z}$,
if $f_i$ calls $f_{a_p}$, then $s_{i,j}$ and $s_{a_p,b_q}$ are in the same order of function call, that is, % or
$s_{i,j}<s_{a_p,b_q}$; otherwise, $s_{i,j}>s_{a_p,b_q}$.
As a running example shown in Fig. \ref{Fig_generating_code_gadgets}, the SeVC is Lines \{7, 9, 10, 11, 13, 14, 16, 18, 22, 23, 24, 25, 26, 1, 3, 4\}, in which the statements in function $func$ appear before the statements in function $println$ because $func$ calls $println$.
The fourth column in Fig. \ref{Fig_generating_code_gadgets} shows the SeVC corresponding to SyVC ``$data$'', namely an order set of
%(not necessarily consecutive)
statements that are semantically related to SyVC ``$data$''.

\subsection{Encoding SeVCs into Vectors}
\label{Design_vector_representation}

Algorithm \ref{alg_obtaining_vectors} encodes SeVCs into vectors in three steps.

\noindent{\bf Step 1 (Lines \ref{algorithm3:begin-step1}-\ref{algorithm3:end-step1} in Algorithm \ref{alg_obtaining_vectors})}.
In order to make SeVCs independent of user-defined variables and function names while capturing program semantic information,
each SeVC $\delta_{i,j,z}$ is transformed to a {\em symbolic representation}. For this purpose, we propose removing non-ASCII characters and comments, then map user-defined variable names to symbolic names (e.g., ``{\sf V1}'', ``{\sf V2}'') in a one-to-one fashion, and finally map user-defined function names to symbolic names (e.g., ``{\sf F1}'', ``{\sf F2}'') in a one-to-one fashion.
% \cite{vuldeepecker}.
Note that different SeVCs may have the same symbolic representation.
%s of different SeVCs may have same symbolic names.
Please refer to \cite{vuldeepecker} for more details about the mapping process.

\noindent{\bf Step 2 (Lines \ref{algorithm3:begin-step2}-\ref{algorithm3:end-step2} in Algorithm \ref{alg_obtaining_vectors})}.
This step is to encode the symbolic representations into vectors. For this purpose, we propose dividing the symbolic representation of a SeVC $\delta_{i,j,z}$ (e.g., ``{\sf V1=V2-8;}'') into a sequence of symbols via a lexical analysis (e.g., ``{\sf V1}'', ``{\sf =}'', ``{\sf V2}'', ``{\sf -}'', ``{\sf 8}'', and ``{\sf ;}'').
%Then the tokens from all symbolic representations yield a large corpus of tokens.
We transform a symbol to a fixed-length vector.
%We adopt a widely used tool $word2vec$ \cite{word2vec} to transform each token in a symbolic representation to a fixed-length vector.
By concatenating the vectors, we obtain a vector $R_{i,j,z}$ for each SeVC.

\noindent{\bf Step 3 (Lines \ref{algorithm3:begin-step3}-\ref{algorithm3:end-step3} in Algorithm \ref{alg_obtaining_vectors})}.
Because (i) the number of symbols (i.e., the vectors representing SeVCs) may be different and (ii) neural networks take vectors of the same length as input, we use a threshold $\theta$ as the length of vectors for the input to neural network.
When a vector is shorter than $\theta$, zeroes are padded to the end of the vector.
When a vector is longer than $\theta$, there are three scenarios but the basic idea is to make the SyVC appear in the middle of the resulting vector.
(i) The sub-vector up to the SyVC is shorter than $\theta/2$. In this case, we delete the rightmost portion of $R_{i,j,z}$ to make the resulting vector have length $\theta$.
(ii) The sub-vector next to the SyVC is shorter than $\theta/2$.
In this case, we delete the leftmost portion of $R_{i,j,z}$ to make the resulting vector have length $\theta$.
(iii) Otherwise, we keep the sub-vector of length $\lfloor (\theta-1)/2 \rfloor$ immediately left to the SyVC and the sub-vector of length $\lceil (\theta-1)/2 \rceil$ immediately right to the SyVC.
Together with the SyVC, we obtain a vector of length $\theta$.
For example, suppose $\theta=15,000$ and the length of each symbol is 30, meaning that each SeVC has 500 symbols. Suppose the number of symbols in a SeVC is 510 (and thus needs to be reduced to 500) and the SyVC is at the position of the 255th symbol (among the 510 symbols), then we retain 249 consecutive symbols immediately left to the SyVC and 250 symbols immediately right to the SyVC. Together with the SyVC, we obtain a vector of 500=249+1+250 symbols. We stress that the preceding operations are well defined because each SyVC is transformed to a SeVC and appears exactly once in the SeVC.

\begin{algorithm}[htb!]
%\algsetup{linenosize=\scriptsize}
\footnotesize
\caption{Encoding SeVCs into vectors}
\label{alg_obtaining_vectors}
\begin{basedescript}{\desclabelstyle{\pushlabel}\desclabelwidth{6em}}
\item[Input:]
A set $Y$ of SyVCs generated by Algorithm \ref{alg_obtaining_SyVC};\\
a set $C$ of SeVCs corresponding to $Y$ and generated by Algorithm \ref{alg_obtaining_ SeVC};\\ 
a threshold $\theta$
\item[Output:]
The set of vectors corresponding to SeVCs
\end{basedescript}
\begin{algorithmic}[1]
\STATE $R \leftarrow \emptyset$;
%\STATE A set of tokens $T \leftarrow \emptyset$;
\FOR{each $\delta_{i,j,z} \in C$ (corresponding to $e_{i,j,z} \in Y$)}  \label{algorithm3:begin-step1}
    \STATE Remove non-ASCII characters in $\delta_{i,j,z}$;
    \STATE Map variable names in $\delta_{i,j,z}$ to symbolic names;
    \STATE Map function names in $\delta_{i,j,z}$ to symbolic names;
\ENDFOR   \label{algorithm3:end-step1}
%\FOR{each $t_j \in T$}
%    \STATE Transform $t_j$ to a fixed-length vector $v_j$;
%\ENDFOR
\FOR{each $\delta_{i,j,z} \in C$ (corresponding to $e_{i,j,z} \in Y$)}
    \STATE $R_{i,j,z} \leftarrow \emptyset$; \label{algorithm3:begin-step2}
    \STATE Divide $\delta_{i,j,z}$ into a set of symbols $S$;
    \FOR{each $\alpha \in S$ in order}
       \STATE Transform $\alpha$ to a fixed-length vector $v(\alpha)$;
       \STATE $R_{i,j,z} \leftarrow R_{i,j,z} || v(\alpha)$, where $||$ means concatenation;
    \ENDFOR   \label{algorithm3:end-step2}
    \IF{$R_{i,j,z}$ is shorter than $\theta$}  \label{algorithm3:begin-step3}
        \STATE Zeroes are padded to the end of $R_{i,j,z}$;
    \ELSIF{the sub-vector (of $\delta_{i,j,z}$) up to the position of the SyVC $e_{i,j,z}$ is shorter than $\theta/2$}
%    \ELSIF{the vector corresponding to $\fs_{i,j}$ is shorter than $\theta/2$}
        \STATE Delete the rightmost portion of $R_{i,j,z}$ to make the resulting vector of length $\theta$;
    \ELSIF{the sub-vector (of $\delta_{i,j,z}$) next to the the position of the SyVC $e_{i,j,z}$ is shorter than $\theta/2$}
%    \ELSIF{the vector corresponding to $\bs_{i,j}$ is shorter than $\theta/2$}
        \STATE Delete the leftmost portion of $R_{i,j,z}$ to make the resulting vector of length $\theta$;
    \ELSE
        \STATE Keep the sub-vector (in $\delta_{i,j,z}$) immediately left to the position of the SyVC of length $\lfloor (\theta-1)/2 \rfloor$, the sub-vector corresponding to the SyVC, and the sub-vector immediately right to the position of the SyVC of length  $\lceil (\theta-1)/2 \rceil$~~\COMMENT{the resulting vector has length $\theta$;}
\ENDIF  \label{algorithm3:end-step3}
    \STATE $R \leftarrow$ $R \cup R_{i,j,z}$;
\ENDFOR
\RETURN $R$; \{the set of vectors corresponding to SeVCs\}
\end{algorithmic}
\end{algorithm}

\subsection{Labeling SeVCs and Corresponding Vectors}
In order to learn a deep neural network, we label the vectors (i.e., the SeVCs they represent) as vulnerable or not as follows:
A SeVC (i.e., the vector representing it) containing a known vulnerability is labeled as ``1'' (i.e., vulnerable), and ``0'' otherwise (i.e., not vulnerable).
A learned deep neural network encodes vulnerability patterns and can detect whether given SeVCs are vulnerable or not.

\section{Experiments and Results}
\label{sec:Experimental_results}
\subsection{Research Questions and Dataset}

\noindent{\bf Research questions}.
Our experiments are geared towards answering the following Research Questions (RQs):
\begin{itemize}
  \item RQ1: Can SySeVR make BLSTM detect multiple kinds (vs. single kind) of vulnerabilities?
  \item RQ2: Can SySeVR make multiple kinds of neural networks to detect multiple kinds of vulnerabilities? Can we explain their (in)effectiveness?
  \item RQ3: Can accommodating control-dependency make SySeVR more effective, and by how much?
  \item RQ4: How more effective are SySeVR-based methods when compared with the state-of-the-art methods?
\end{itemize}
In order to answer these questions, we implement the deep neural networks in Python using Tensorflow \cite{DBLP:conf/osdi/AbadiBCCDDDGIIK16}.
The computer running experiments has a NVIDIA GeForce GTX 1080 GPU and an Intel Xeon E5-1620 CPU running at 3.50GHz.

\noindent{\bf Vulnerability dataset.}
We produce a vulnerability dataset from two sources: NVD \cite{NVD} and SARD \cite{SARD}.
NVD contains vulnerabilities in software products (i.e., software systems) and possibly {\tt diff} files describing the difference between a vulnerable piece of code and its patched version. SARD contains production, synthetic and academic programs (also known as {\em test cases}), which are categorized as ``good'' (i.e., having no vulnerabilities), ``bad'' (i.e., having vulnerabilities), and ``mixed'' (i.e., having vulnerabilities whose patched versions are also available).
Note that a program in NVD consists of one or several files (e.g., .c or .cpp files) which contain some vulnerability (corresponding to a CVE ID) or its patched version,
and that a program in SARD is a test case.

For NVD, we focus on 19 popular C/C++ open source products (same as in \cite{vuldeepecker}) and their vulnerabilities that are accompanied by {\tt diff} files, which are needed for extracting vulnerable pieces of code.
%the syntax characteristics of vulnerabilities}.
As a result, we collect 1,591
%1,592
open source C/C++ programs, of which 874 are vulnerable.
For SARD, we collect 14,000 C/C++ programs, of which 13,906 programs are vulnerable (i.e., ``bad'' or ``mixed''). 
Note that a large number of these vulnerable programs belong to the ``mixed'' category and come with both the vulnerable functions and their patched versions.
The average length of these programs is 573.5 lines of code.
%In total, we collect 16,186 programs, of which 14,571 are vulnerable.
In total, we collect 15,591 programs, of which 14,780 are vulnerable; these vulnerable programs contain 126 {\em types} of vulnerabilities, where each type is uniquely identified by a {\em Common Weakness Enumeration IDentifier} (CWE ID) \cite{CWE}.
%at the 3rd layer of the CWE ID tree in the {\em research concept} view \cite{CWE}.
The 126 CWE IDs are published with our dataset.

\subsection{Evaluation Metrics}

The effectiveness of vulnerability detectors can be evaluated by the following widely-used metrics \cite{DBLP:journals/csur/PendletonGCX17}: {\em false-positive rate} ($FPR$), {\em false-negative rate} ($FNR$), {\em accuracy} ($A$), {\em precision} ($P$),
%{\em recall} ($R$),
{\em F1-measure} ($F1$), and {\em Matthews Correlation Coefficient} (MCC) \cite{MATTHEWS1975442}.
Let {\sf TP} denote the number of vulnerable samples that are detected as vulnerable, {\sf FP} denote the number of samples are not vulnerable but are detected as vulnerable, {\sf TN} denote the number of samples that are not vulnerable (dubbed {\em non-vulnerable}) and are detected as not vulnerable, and {\sf FN} denote the number of vulnerable samples that are detected as not vulnerable.
The metric $FPR=\frac{{\sf FP}}{{\sf FP}+{\sf TN}}$ measures the proportion of false-positive samples among the samples that are not vulnerable.
The metric $FNR=\frac{{\sf FN}}{{\sf TP}+{\sf FN}}$ measures the proportion of false-negative samples among the vulnerable samples.
The metric $A=\frac{{\sf TP}+{\sf TN}}{{\sf TP}+{\sf FP}+{\sf TN}+{\sf FN}}$ measures the proportion of correctly detected samples among all samples.
The metric $P=\frac{{\sf TP}}{{\sf TP}+{\sf FP}}$ measures the proportion of truly vulnerable samples among the detected (or claimed) vulnerable samples.
The metric $F1=\frac{2 \cdot P \cdot (1-FNR) }{P + (1-FNR)}$ measures the overall effectiveness by considering both precision and false-negative rate.
The metric $MCC=\frac{{\sf TP} \times {\sf TN} - {\sf FP} \times {\sf FN}}{\sqrt{({\sf TP}+{\sf FP})({\sf TP}+{\sf FN})({\sf TN}+{\sf FP})({\sf TN}+{\sf FN})}}$
measures the degree to which model predictions match ground-truth labels; this metric is useful especially when dealing with imbalanced data, which is the case of the present paper because we have many more non-vulnerable samples than vulnerable ones.

\subsection{Experiments}
\label{subsec:Preparing_data}

The experiments follow the SySeVR framework, with elaborations when necessary.

\subsubsection{Extracting SyVCs}
\label{subsubsection:extracting-SyVCs}

In what follows we will elaborate the two components in Algorithm \ref{alg_obtaining_SyVC} that are specific to different kinds of vulnerabilities: the extraction of vulnerability syntax characteristics and how to match them.

\noindent{\bf Extracting vulnerability syntax characteristics.}
In order to extract syntax characteristics of known vulnerabilities, it would be natural to extract the vulnerable lines of code from the vulnerable programs mentioned above, and analyze their syntax characteristics. However, this is an extremely time-consuming task, which prompts us to leverage the C/C++ vulnerability rules of a state-of-the-art commercial tool, Checkmarx \cite{Checkmarx}, to analyze vulnerability syntax characteristics.
As we will see, this alternate method is effective because it covers 93.6\% of the vulnerable programs collected from SARD.
It is worth mentioning that we choose Checkmarx over open-source tools (e.g., Flawfinder \cite{FlawFinder} and RATS \cite{RATS}) because the latter have %weaknesses
simple parsers and imperfect rules \cite{DBLP:phd/dnb/Yamaguchi15}.

\begin{figure}[!htbp]
\vspace{-0.3cm}
\centering
\includegraphics[width=.23\textwidth]{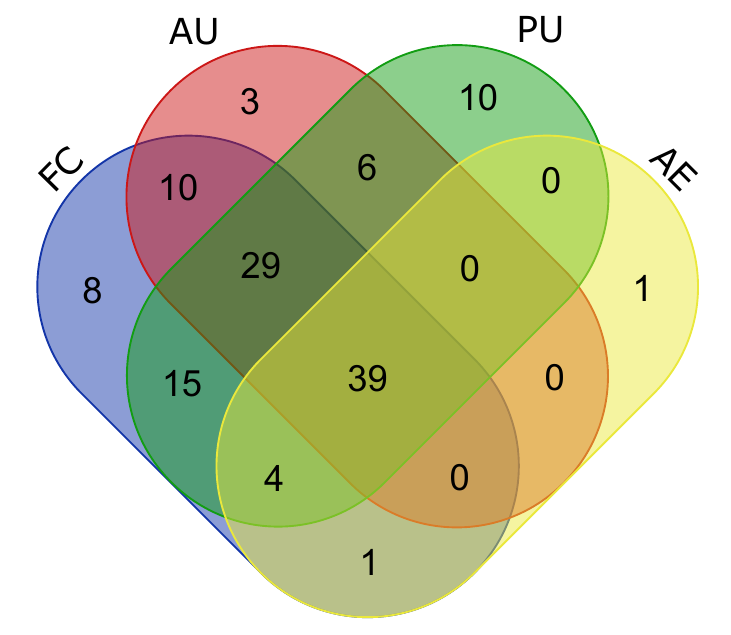}
\vspace{-0.2cm}
\caption{Venn diagram of the FC, AU, PU, and AE in terms of the CWE IDs they cover, where $|\text{FC}|=106$, $|\text{AU}|=87$, $|\text{PU}|=103$, $|\text{AE}|=45$, and $|\text{FC} \cup \text{AU}\cup \text{PU}\cup \text{AE}|=126$.}
%, and a number is the number of CWE IDs corresponding to an intersection.
\vspace{-0.2cm}
\label{Fig_venn}
\end{figure}

Our {\em manual} examination of Checkmarx rules leads to the following 4 kinds of vulnerability syntax characteristics (each accommodating many vulnerabilities).
%\vspace{-0.1cm}
\begin{itemize}
\item {\em Library/API Function Call} (FC for short):
This kind of syntax characteristic covers 811 library/API function calls, which are published with our dataset.
%which are listed in Table \ref{Table_sensitive_function} of Appendix \ref{sec:appendix_FC}.
These 811 function calls correspond to 106 CWE IDs.

\item {\em Array Usage} (AU for short): This kind of syntax characteristic covers 87 CWE IDs related to arrays (e.g., issues related to array element access, array address arithmetic).

\item {\em Pointer Usage} (PU for short): This kind of syntax characteristic covers 103 CWE IDs related to pointers (e.g., improper use in pointer arithmetic, reference, address transfer as a function parameter).

\item {\em Arithmetic Expression} (AE for short): This kind of syntax characteristic covers 45 CWE IDs related to improper arithmetic expressions (e.g., integer overflow).
\end{itemize}
Fig. \ref{Fig_venn} shows that these 4 kinds of syntax characteristics overlap with each other in terms of the CWE IDs they cover.
These 4 kinds of syntax characteristics are generated from programs corresponding to 126 CWE IDs.
Note that one kind of syntax characteristics may cover multiple CWE IDs and that one CWE ID may be covered by one or multiple kinds of syntax characteristics.
For example, Fig. \ref{Fig_venn} shows that the vulnerabilities corresponding to 10 CWE IDs are covered by the PU-kind syntax characteristics but not others, and the vulnerabilities corresponding to 39 CWE IDs are covered by all of the 4 kinds of syntax characteristics (i.e., FC, AU, PU, and AE).

\begin{figure}[!htbp]
\vspace{-0.2cm}
\centering
\includegraphics[width=.45\textwidth]{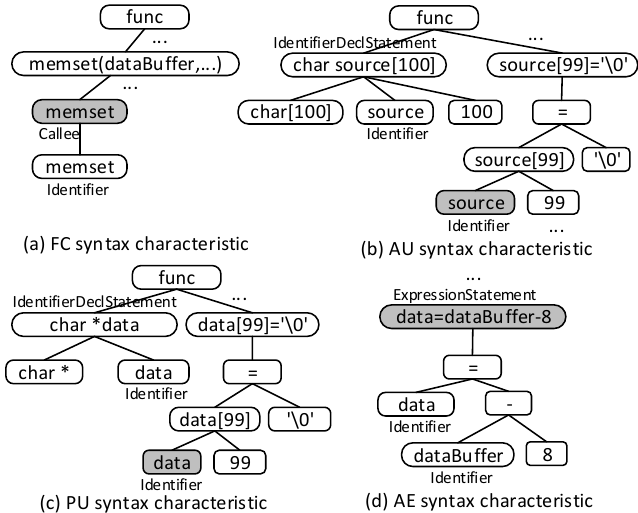}
\vspace{-0.2cm}
\caption{Examples for illustrating the matching of syntax characteristics, where a highlighted node matches some vulnerability syntax characteristic and therefore is a SyVC.
}
\vspace{-0.2cm}
\label{Fig_syntax_characteristics}
\end{figure}

\noindent{\bf Matching syntax characteristics.}
In order to use Algorithm \ref{alg_obtaining_SyVC} to extract SyVCs, we need to determine whether or not a code element $e_{i,j,z}$, which is on the abstract syntax tree $T_i$ of function $f_i$ in program $P$,
%$P=\{f_1,f_2,\ldots,f_\eta\}$,
matches a vulnerability syntax characteristic. %mentioned above.
Note that $T_i$ can be generated by using {\em Joern} \cite{yamaguchi2014modeling}.
The following method, as illustrated in Fig. \ref{Fig_syntax_characteristics} via the example program shown in Fig. \ref{Fig_three_representation},
can automatically decide whether or not code element $e_{i,j,z}$ matches a syntax characteristic.

\begin{itemize}
\item As illustrated in Fig. \ref{Fig_syntax_characteristics}(a), we say code element $e_{i,j,z}$ (i.e., ``memset'') matches the FC syntax characteristic  if (i) $e_{i,j,z}$ on $T_i$ is a ``callee'' (i.e., the function is called), and (ii) $e_{i,j,z}$ is one of the 811 function calls mentioned above.

\item As illustrated in Fig. \ref{Fig_syntax_characteristics}(b), we say code element $e_{i,j,z}$ (i.e., ``source'') matches the AU syntax characteristic if
     (i) $e_{i,j,z}$ is an identifier declared in an identifier declaration statement (i.e., IdentifierDeclStatement) node and (ii) the IdentifierDeclStatement node contains characters `[' and `]'.

\item As illustrated in Fig. \ref{Fig_syntax_characteristics}(c), we say code element $e_{i,j,z}$ (i.e., ``data'') matches the PU syntax characteristic if
     (i) $e_{i,j,z}$ is an identifier declared in an IdentifierDeclStatement node and (ii) the IdentifierDeclStatement node contains character `$\ast$'.

\item As illustrated in Fig. \ref{Fig_syntax_characteristics}(d), we say code element $e_{i,j,z}$ (``data=dataBuffer-8'') matches the AE syntax characteristic if (i) $e_{i,j,z}$ is an expression statement (ExpressionStatement) node and (ii)  $e_{i,j,z}$ contains a character `=' and has one or more identifiers on the right-hand side of `='.
%, then the code element $e_{i,j,z}$ satisfies the syntax characteristics of AE.
\end{itemize}

\noindent{\bf Extracting SyVCs.}
Now we can use Algorithm \ref{alg_obtaining_SyVC} to extract SyVCs from the 15,591 programs.
Corresponding to the 4 kinds of syntax characteristics, we extract 4 kinds of SyVCs:
\begin{itemize}
\item FC-kind SyVCs: We extract 6,356 from NVD and 58,047 from SARD, or 64,403 in total.

\item AU-kind SyVCs: We extract 9,812 from NVD and 32,417 from SARD, or 42,229 in total.

\item PU-kind SyVCs: We extract 73,890 from NVD and 217,951 from SARD, or 291,841 in total.

\item AE-kind SyVCs: We extract 5,295 from NVD and 16,859 from SARD, or 22,154 in total.
% vulnerabilities.% (CWE-190).
\end{itemize}
Putting them together,
we extract 420,627 SyVCs, which cover 13,016 (out of the 13,906, or 93.6\%) vulnerable programs collected from SARD; this coverage validates our idea of using Checkmarx rules to derive vulnerability syntax characteristics. Note that we can compute the coverage 93.6\% because SARD gives the precise location of each vulnerability; in contrast, we cannot compute the coverage with respect to NVD because it does not give precise locations of vulnerabilities.
The average time for extracting a SyVC is 270 milliseconds.

\subsubsection{Transforming SyVCs to SeVCs}

When using Algorithm \ref{alg_obtaining_ SeVC} to transform SyVCs to SeVCs, we use {\em Joern} \cite{yamaguchi2014modeling} to extract PDGs.
% of the 16,186 programs mentioned above.
Corresponding to the 420,627 SyVCs extracted from Algorithm \ref{alg_obtaining_SyVC}, Algorithm \ref{alg_obtaining_ SeVC} generates 420,627 SeVCs (while recalling that one SyVC is transformed to one SeVC). In order to see the effect of semantic information, we actually use Algorithm \ref{alg_obtaining_ SeVC} to generate two sets of SeVCs: one set accommodating semantic information induced by data dependency only, and the other set accommodating semantic information induced by both data dependency and control dependency. In either case, the second column of Table \ref{Table_dataset} summarizes the numbers of SeVCs categorized by the kinds of SyVCs from which they are transformed.
In terms of the efficiency of the SyVC$\to$SeVC transformation, on average it takes 331 milliseconds to generate a SeVC accommodating data dependency
and 362 milliseconds to generate a SeVC accommodating data dependency and control dependency.

\begin{table}[!htbp]
\vspace{-0.2cm}
\caption{\small The number of SeVCs, vulnerable SeVCs, and non-vulnerable SeVCs from the 15,591 programs}
\vspace{-0.2cm}
\label{Table_dataset}
\centering
\footnotesize
\begin{tabular}{|c|r|r|r|}
\hline
Kind of SyVCs & \tabincell{c}{\#SeVCs } & \tabincell{c}{\#Vul. SeVCs} & \tabincell{c}{\#Non-vul. SeVCs}\\
\hline
FC-kind & 64,403 & 13,603 & 50,800\\
\hline
AU-kind & 42,229 & 10,926 & 31,303 \\
\hline
PU-kind & 291,841 & 28,391 & 263,450 \\
\hline
AE-kind & 22,154 & 3,475 & 18,679 \\
\hline
Total & 420,627 & 56,395 & 364,232  \\
\hline
\end{tabular}
\vspace{-0.2cm}
\end{table}

\subsubsection{Encoding SeVCs into Vector Representation}
%As described in Section \ref{Design_vector_representation},
We use Algorithm \ref{alg_obtaining_vectors} to encode SeVCs into vectors.
For this purpose, we adopt $word2vec$ \cite{word2vec} to encode the symbols
% (as in Definition \ref{definition:program})
of the SeVCs (extracted from the 15,591 programs) into fixed-length vectors. The main hyper-parameters include: the dimensionality of word vectors is 30, the window size is 5, the training algorithm is skip-gram, and the threshold for configuring which higher-frequency words are randomly downsampled is 0.001. 
Then, each SeVC is represented by the concatenation of the vectors representing its symbols.
We set each SeVC to have 500 symbols (padding or truncating if necessary, as discussed in Algorithm \ref{alg_obtaining_vectors}) and the length of each symbol is 30, meaning $\theta=15,000$.

\subsubsection{Generating Ground-truth Labels of SeVCs}
We generate ground-truth labels for the SeVCs in two steps. 
First, we generate {\em preliminary} labels automatically.
For SeVCs extracted from NVD, we examine the vulnerabilities whose {\tt diff} files contain {\em line deletion}, while noting that we do not consider the {\tt diff} files that only contain {\em line addition} because NVD does not give the vulnerable statements in such cases.
%We generate the ground truth labels automatically in the following two steps.
For a {\tt diff} file containing {\em line deletion}, we parse it to mark and distinguish (i) the lines (i.e., statements) that are prefixed with ``-'' and are deleted/modified from (ii) the lines that are prefixed with ``-'' and are moved (i.e., deleted at one place and added at another place).
If a SeVC contains at least one deleted/modified statement that is prefixed with ``-'', it is labeled as ``1'' (i.e., vulnerable); if a SeVC contains at least one moved statement prefixed with ``-'' and the detected file contains a known vulnerability, it is labeled as ``1''; otherwise, it is labeled as ``0'' (i.e., not vulnerable).
For SeVCs extracted from SARD, a SeVC extracted from a ``good'' program is labeled as ``0''  (i.e., not vulnerable); a SeVC extracted from a ``bad'' or ``mixed'' program is labeled as ``1''  (i.e., vulnerable) if the SeVC contains at least one vulnerable statement; otherwise, it is labelled as ``0''.

Second, in order to improve the quality of the preliminary labels mentioned above, we use stratified $k$-fold ($k$=5) cross validation to identify the vulnerable SeVCs that may have been mislabeled in the previous step (while noting that a true vulnerable sample is never mislabelled as ``0'')
%with high probability 
and check them manually, as follows. (i) The dataset 
%\footnote{double check to use "dataset" or "data set" throughout the paper} 
is divided into 5 subsets. (ii) One subset is used as the validation set and the other 4 subsets are put together as the training set. (iii) The samples in the validation set are classified by the trained neural network. The false-negatives (i.e., the vulnerable samples that are not detected as vulnerable) are considered as the samples that may have been mislabeled.
%with high probability. 
Then, we manually check these samples and correct the mislabeled samples. Steps (ii) and (iii) are repeated 5 times such that each subset is used as the validation set once.
In total, we manually check the 2,605 samples that may have been mislabeled (i.e., 0.6\% of all 420,627 samples). Among these 2,605 samples, we manually corrected 1,641 false-negatives (while noting that there are no false-positives because these 2,605 samples are all vulnerable).

In total, 56,395 SeVCs are labeled as ``1'' and 364,232 SeVCs are labeled as ``0''. The third and fourth columns of Table \ref{Table_dataset} summarize the number of vulnerable vs. not vulnerable SeVCs corresponding to each kind of SyVCs.
The ground-truth label of the vector corresponding to a SeVC is the same as the ground-truth label of the SeVC.

\subsection{Experimental Results}
For the programs collected from NVD and SARD, we  
%shuffle the programs and 
randomly select 80\% of them as the training set
(i.e., for training and validation) and the rest 20\% of programs as the test set (i.e., for testing), respectively.

\subsubsection{Experiments for Answering RQ1}
\label{Experiment_RQ1}

In this experiment, we use BLSTM as in \cite{vuldeepecker} and the SeVCs accommodating semantic information induced by data and control dependencies.
We randomly choose 30,000 SeVCs extracted from the training programs as the training set and 7,500 SeVCs extracted from the test programs as the test set.
Both sets contain SeVCs corresponding to the 4 kinds of SyVCs, proportional to the ratio of vulnerable vs. non-vulnerable SeVCs in each kind of SyVCs.
%as shown in the third and fourth columns of Table \ref{Table_dataset}.
For fair comparison with VulDeePecker \cite{vuldeepecker}, we also randomly choose 30,000 SeVCs corresponding to the FC-kind SyVCs extracted from the training programs as the training set, and 7,500 SeVCs corresponding to the FC-kind SyVCs extracted from the test programs as the test set (also proportional to the ratio of vulnerable vs. non-vulnerable SeVCs in the entire set of FC-kind SyVCs). Note that these SeVCs only accommodate semantic information induced by data dependency (as in \cite{vuldeepecker}).
We use the stratified 5-fold cross-validation to train deep neural networks, and choose the values of hyper-parameters that lead to the highest F1-measure (i.e., the overall vulnerability detection effectiveness).
The main hyper-parameters we use to learn BLSTM are described as follows. The dropout is 0.2; the batch size is 16; the number of epochs is 20; the output dimension is 256; the minibatch stochastic gradient descent together with ADAMAX \cite{kingma2014adam} is used for training with a default learning rate of 0.002; the dimension of hidden vectors is 500; and the number of hidden layers is 2.

\begin{table}[!htbp]
%\vspace{-0.2cm}
\caption{\small Effectiveness of VulDeePecker \cite{vuldeepecker} vs. SySeVR-enabled BLSTM (or SySeVR-BLSTM) for detecting vulnerabilities related to various kinds of SyVCs (metrics unit: \%)}
\vspace{-0.2cm}
\label{Table_BLSTM_four_types_SyVCs}
\centering
%\scriptsize
\footnotesize
\begin{tabular}{|c|c|c|c|c|c|c|c|}
\hline
\tabincell{c}{Method} & \tabincell{c}{Kind of\\ SyVC} &  \tabincell{c}{FPR} &  \tabincell{c}{FNR} &  \tabincell{c}{A} &  \tabincell{c}{P}  &  \tabincell{c}{F1} & MCC \\
\hline
\tabincell{c}{VulDee-\\Pecker} & FC-kind & 5.5 & 22.5 & 90.8 & 79.1 & 78.3 & 72.5\\
\hline
\hline
{\multirow{5}{*}{\tabincell{c}{SySeVR-\\BLSTM}}} & FC-kind & 2.1 & 17.5 & 94.7 & 91.5 & 86.8  & 83.6\\
\cline{2-8}
 & AU-kind & 3.8 & 17.1 & 92.7 & 88.3 & 85.5  & 80.7\\
\cline{2-8}
 & PU-kind & 1.3 & 19.7 & 96.9 & 87.3 & 83.7  & 82.1\\
\cline{2-8}
 & AE-kind & 1.5 & 18.3 & 96.6 & 87.9 & 84.7  & 82.9\\
\cline{2-8}
 & All-kinds & 1.7 & 19.0 & 96.0 & 88.0 & 84.4  & 82.2\\
\hline
\end{tabular}
\vspace{-0.2cm}
\end{table}

Table \ref{Table_BLSTM_four_types_SyVCs} summarizes the results.
We observe that SySeVR-BLSTM can detect vulnerabilities of the AU-kind with the lowest FNR (17.1\%), but with a higher FPR than the other three kinds of vulnerabilities. It detects vulnerabilities of the FC-kind with the highest F1-measure (86.8\%) and MCC (83.6\%). The other three kinds of vulnerabilities lead to, on average, a FPR of 1.6\% and a FNR of 18.5\%.
Overall, SySeVR-BLSTM achieves a 3.4\% lower FPR and a 5.0\% lower FNR than VulDeePecker when applied to detect vulnerabilities of the same kind (i.e., the FC-kind). 
This can be explained by the fact that SySeVR-BLSTM 
%involves more kinds of vulnerabilities and 
accommodates more semantic information (e.g., control dependency) via SeVCs.
This leads to:

\begin{insight}
{\em SySeVR-BLSTM can detect vulnerabilities related to function calls, array usage, pointer usage and arithmetic expressions, and can achieve a 3.4\% lower FPR and a 5.0\% lower FNR when compared with VulDeePecker in detecting vulnerabilities related to library/API function calls.}
\end{insight}

\subsubsection{Experiments for Answering RQ2}
\label{Experiment_RQ2}
In order to answer RQ2, we use the stratified 5-fold cross-validation to train 8 standard models: a linear {\em Logistic Regression} (LR) classifier, a neural network with one hidden layer {\em Multi-Layer Perception} (MLP), a DBN \cite{DBLP:journals/scholarpedia/Hinton09}, a CNN \cite{DBLP:journals/tnn/LawrenceGTB97}, and four RNNs (i.e., {\em Long Short-Term Memory} (LSTM), {\em Gated Recurrent Unit} (GRU), BLSTM, and BGRU \cite{hochreiter1997long, DBLP:conf/ssst/ChoMBB14, DBLP:journals/nn/GravesS05}), using the same dataset (of 4 kinds of SyVCs) as in Section \ref{Experiment_RQ1}.
In each case, we choose the hyper-parameter value that leads to the highest F1-measure.

\begin{table}[!htbp]
\vspace{-0.2cm}
\caption{\small Effectiveness of SySeVR-enabled different kinds of models in detecting the 4 kinds of vulnerabilities (metrics unit: \%)}
% for detecting 143 types of vulnerabilities corresponding to 4 kinds of SyVCs (i.e., 4 kinds of SeVCs)}
%when trained from, and applied to, all of the 4 classes of vulnerabilities.}
\vspace{-0.2cm}
\label{Table_neural_networks_four_types_SyVCs_FPR0.2}
\centering
%\scriptsize
\footnotesize
\begin{tabular}{|c|c|c|c|c|c|c|}
\hline
 \tabincell{c}{Model} &  \tabincell{c}{FPR} &  \tabincell{c}{FNR} &  \tabincell{c}{A}&  \tabincell{c}{P}  &  \tabincell{c}{F1} &  \tabincell{c}{MCC}\\
\hline
LR & 2.0 & 45.5 & 92.1 & 80.8 & 65.1 & 62.5 \\
\hline
MLP & 2.0 & 37.3 & 93.1 & 82.1 & 71.1 & 68.1 \\
\hline
DBN & 2.0 & 44.0 & 91.6 & 82.1 & 66.6 & 63.5 \\
\hline
CNN & 2.0 & 17.9 & 95.7 & 85.6 & 83.8 & 81.4 \\
\hline
LSTM & 2.0 & 21.7 & 95.2 & 85.2 & 81.6 & 79.0 \\
\hline
GRU & 2.0 & 17.6 & 95.7 & 85.7 & 84.0 & 81.7\\
\hline
BLSTM & 2.0 & 15.7 & 96.0 & 86.2 & 84.3 & 83.0 \\
\hline
BGRU & 2.0 & 14.7 & 96.0 & 86.4 & 85.8 & 83.7\\
\hline
\end{tabular}
%\vspace{-0.2cm}
\end{table}

Table \ref{Table_neural_networks_four_types_SyVCs_FPR0.2} summarizes the results by setting FPR to 2.0\% in each model, which is chosen
%. We choose 2.0\% as the FPR to compare the models
because it is the FPR of the model that achieves the highest F1-measure.
We observe that when compared with unidirectional RNNs (i.e., LSTM and GRU), bidirectional RNNs (i.e., BLSTM and BGRU) can respectively improve the FNR by 4.5\% and the F1-measure by 2.3\% on average. This improvement might be caused by the following: Bidirectional RNNs can accommodate more information about the statements that appear before and after the statement in question.
We further observe that bidirectional RNNs (especially BGRU) are more effective than CNN, which in turn is more effective than DBN and shallow learning models (i.e., LR and MLP).
%Considering the imbalance in the data set, The MCC metric shows that the effect of imbalanced class size in the dataset on the results.
Moreover, these models achieve a similar effectiveness in both MCC and F1-measure, meaning that the issue of data imbalance is not significant. 
In summary,
\begin{insight}
%For SySeVR-based vulnerability detection,
{\em SySeVR-enabled bidirectional RNNs (especially BGRU) are more effective than SySeVR-enabled unidirectional RNNs and CNN, which are more effective than SySeVR-enabled DBN and shallow learning models (i.e., LR and MLP).
Still, FNRs of all these models are consistently much higher than their FPRs.}
\end{insight}

SySeVR-enabled models mentioned above adopt word2vec \cite{word2vec} to generate vectors.
In order to see if word2vec can be replaced by a simpler vector representation, say {\em token frequency}, we use bag-of-words \cite{DBLP:journals/mlc/ZhangJZ10} to encode SeVCs into fixed-length vectors. With this vector representation, we conduct experiments using two shallow models (i.e., LR and MLP) and two deep neural networks (i.e., CNN and BGRU). Table \ref{Table_neural_networks_four_types_BoW} reports the experimental results.
We observe that the best result for word2vec (BGRU achieving an F1 of 85.8\% and a MCC of 83.7\% as shown in Table \ref{Table_neural_networks_four_types_SyVCs_FPR0.2}) is much better than the best result for bag-of-words (MLP achieving an F1 of 76.6\% and a MCC of 73.7\%).
For bag-of-words, we observe that shallow models are more effective than deep neural networks; for word2vec, deep neural networks are more effective than shallow models.
In particular, BGRU, which is the most effective for word2vec (F1 of 85.8\% and MCC of 83.7\%), is the least effective for bag-of-words (F1 of 48.8\% and MCC of 46.9\%). This can be explained by the fact that there is no context information for the vectors generated by bag-of-words, causing BGRU not to be able to capture the context and achieve a low effectiveness. This leads to:
\begin{insight}
{\em Using a distributed representation, such as word2vec, to capture context information is important to SySeVR. In particular, a representation centered at token frequency is not sufficient.} 
\end{insight}
Because of this, we always use word2vec to generate vectors for the experiments that will be discussed in the rest of the paper.

\begin{table}[!htbp]
\vspace{-0.2cm}
\caption{\small Effectiveness of SySeVR-enabled models using vectors derived from bag-of-words (metrics unit: \%)}
\vspace{-0.2cm}
\label{Table_neural_networks_four_types_BoW}
\centering
%\scriptsize
\footnotesize
\begin{tabular}{|c|c|c|c|c|c|c|}
\hline
 \tabincell{c}{Model} &  \tabincell{c}{FPR} &  \tabincell{c}{FNR} &  \tabincell{c}{A}&  \tabincell{c}{P}  &  \tabincell{c}{F1} &  \tabincell{c}{MCC}\\
\hline
LR & 2.0 & 34.4 & 93.5 & 83.4 & 73.4 & 70.5 \\
\hline
MLP & 2.0 & 29.7 & 94.1 & 84.2 & 76.6 & 73.7 \\
\hline
CNN & 2.0 & 55.1 & 90.6 & 76.8 & 56.7 & 54.2 \\
\hline
BGRU & 2.0 & 63.3 & 89.5 & 72.8 & 48.8 & 46.9\\
\hline
\end{tabular}
%\vspace{-0.2cm}
\end{table}

\noindent{\bf Towards explaining the effectiveness of BGRU in vulnerability detection.}
It is important, but an outstanding open problem, to explain the effectiveness of deep neural networks.
Now we report our initial effort along this direction. In what follows we focus on BGRU because it is more effective than the others.
\ignore{
In order to analyze why the false-negatives are high,
we plot in Fig. \ref{Fig_Distribution_of_CWE_IDs} the distribution of the vulnerabilities (according to their CWE IDs) that are missed by BGRU (i.e., false-negatives). We observe that
the vulnerabilities corresponding to CWE-284 cause the highest FNR (i.e., 50\%), and that the vulnerabilities corresponding to CWE-706 cause the lowest FNR (i.e., 1.0\%).
We find that the vulnerabilities causing a higher FNR have a smaller amount of data in the training set, which means that the amount of training data describing vulnerabilities plays a critical role (i.e., the more vulnerability samples, the fewer false-negatives).
A similar phenomenon is observed for the other neural networks.

\begin{figure}[!htbp]
\vspace{-0.2cm}
\centering
\includegraphics[width=0.38\textwidth]{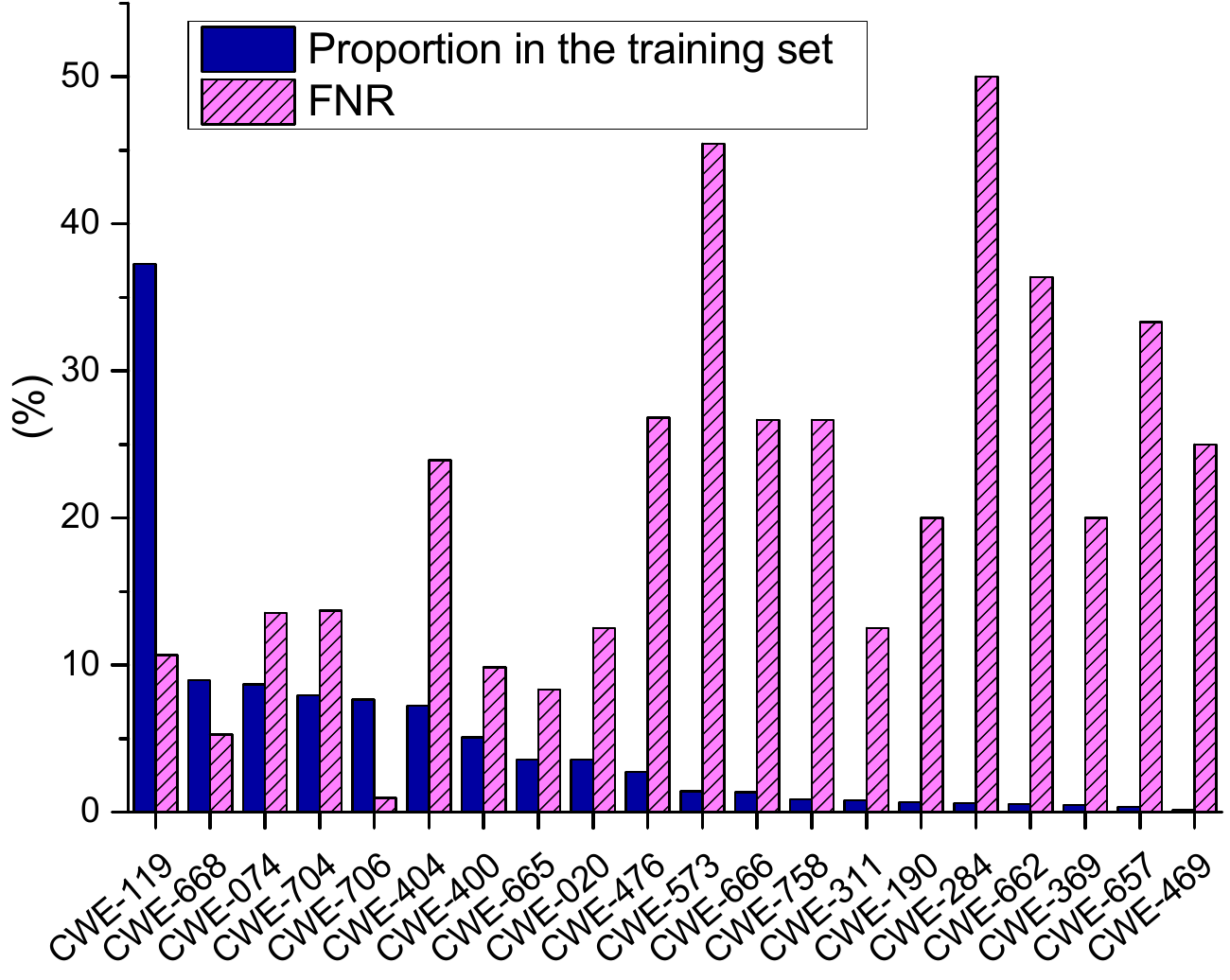}
\vspace{-0.4cm}
\caption{\small{Explaining the cause of BGRU's high FNRs: the smaller the amount of training data for a type of vulnerabilities, the higher the FNR.}}
% The CWE IDs with higher FNR account for lower proportions in the training set.}
\vspace{-0.3cm}
\label{Fig_Distribution_of_CWE_IDs}
\end{figure}
}

\begin{figure}[!htbp]
\vspace{-0.4cm}
\centering
\includegraphics[width=.38\textwidth]{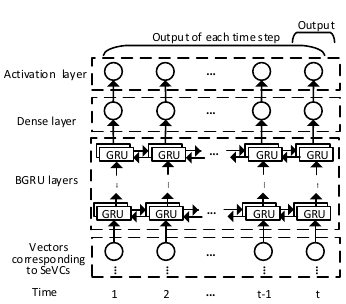}
\vspace{-0.2cm}
\caption{The structure of BGRU}
\label{Fig_bgru}
\vspace{-0.2cm}
\end{figure}

In order to explain the effectiveness of BGRU, we review its structure in Fig. \ref{Fig_bgru}.
For each SeVC and each time step, there is an output (belonging to $[0,1]$) at the activation layer.
The output of BGRU is the output of the last time step at the activation layer; the closer this output is to 1, the more likely the SeVC is classified as vulnerable.
For the classification of a SeVC, we identify the tokens (i.e., the symbols representing them) that play a critical role in determining its classification.
This can be achieved by looking at all %pairs tokens
pairs of tokens at time steps $(t',t'+1)$.
We find that if the activation-layer output corresponding to the token at time step %$t'$
$t'+1$ is substantially (e.g., 0.6) greater (vs. smaller)
than the activation-layer output corresponding to the token at time step
$t'$, then the token at time step %$t'$ ($t'+1$)
$t'+1$ plays a critical role in classifying the SeVC as vulnerable (correspondingly, not vulnerable).
Moreover, we find that some false-negatives are caused by the token ``{\tt if}'' or the tokens following it, because these tokens frequently appear
%often appear 
in SeVCs that are not vulnerable.
We also find that some false-positives are caused by the tokens related to library/API function calls and their arguments, because these tokens frequently appear
%often appear 
in SeVCs that are vulnerable.
In summary,

\begin{insight}
{\em
If a syntax element (e.g., token) appears in vulnerable (resp. non-vulnerable) SeVCs much more frequently than appearing in non-vulnerable (resp. vulnerable) ones, the syntax element may cause false-positives (resp. false-negatives); this means that  appearance frequency of syntax elements matters.
}
\end{insight}

\subsubsection{Experiments for Answering RQ3}
\label{Experiment_RQ3}

We use experiments to compare the effectiveness of (i) the 8 models learned from the SeVCs that accommodate semantic information induced by data dependency and (ii) the 8 models learned from the SeVCs that accommodate semantic information induced by data dependency and control dependency.
In either case, we randomly choose 30,000 SeVCs extracted from the training programs as the training set and 7,500 SeVCs extracted from the test programs as the test set.
All of these training and test sets correspond to the 4 kinds of SyVCs, proportional to the amount of vulnerable vs. non-vulnerable SeVCs for each kind of SyVCs.

\begin{table}[!htbp]
%\renewcommand{\arraystretch}{1.2}
\vspace{-0.2cm}
\caption{\small Effectiveness of semantic information induced by data dependency (``DD'' for short) vs. induced by data dependency and control dependency (``DDCD'' for short) (metrics unit: \%)}
\vspace{-0.2cm}
\label{Table_Comparison_code_gadgets}
\centering
%\scriptsize
\footnotesize
\begin{tabular}{|c|c|c|c|c|c|c|c|}
\hline
 \tabincell{c}{Model} &  \tabincell{c}{Kind of\\ SeVC} &  \tabincell{c}{FPR} &  \tabincell{c}{FNR} &  \tabincell{c}{A}&  \tabincell{c}{P}  &  \tabincell{c}{F1} &  \tabincell{c}{MCC}\\
\hline
{\multirow{2}{*}{LR}} & DD & 2.0 & 69.7 & 88.6 & 69.7 & 42.2 & 41.0\\
\cline{2-8}
 & DDCD & 2.0 & 45.5 & 92.1 & 80.8 & 65.1 & 62.5 \\
 \hline
{\multirow{2}{*}{MLP}} & DD &  2.0 & 66.9 & 89.0 & 72.0 & 45.4  & 44.0 \\
\cline{2-8}
 & DDCD & 2.0 & 37.3 & 93.1 & 82.1 & 71.1 & 68.1 \\
 \hline
{\multirow{2}{*}{DBN}} & DD &  2.0 &  78.5 & 87.4 &  63.0 & 32.0  &  31.7 \\
\cline{2-8}
 & DDCD & 2.0 & 44.0 & 91.6 & 82.1 & 66.6 & 63.5 \\
\hline
{\multirow{2}{*}{CNN}} & DD &  2.0 &  42.9 & 92.3 & 81.3 & 67.0  & 64.0\\
\cline{2-8}
 & DDCD & 2.0 & 17.9 & 95.7 & 85.6 & 83.8 & 81.4 \\
\hline
{\multirow{2}{*}{LSTM}} & DD &  2.0 & 68.6 &  88.7 & 70.0 & 43.4  & 41.9 \\
\cline{2-8}
 & DDCD & 2.0 & 21.7 & 95.2 & 85.2 & 81.6 & 79.0 \\
\hline
{\multirow{2}{*}{GRU}} & DD &  2.0 &  42.8 &  92.3 & 81.7 &  67.3  &  64.4 \\
\cline{2-8}
 & DDCD & 2.0 & 17.6 & 95.7 & 85.7 & 84.0 & 81.7 \\
\hline
{\multirow{2}{*}{BLSTM}} & DD &  2.0 &  45.7 & 92.1 &  82.1 & 65.3  &  62.7 \\
\cline{2-8}
 & DDCD & 2.0 & 15.7 & 96.0 & 86.2 & 84.3 & 83.0 \\
\hline
{\multirow{2}{*}{BGRU}} & DD &  2.0 &  42.3 & 92.5 & 82.3 &  67.8  &  65.0 \\
\cline{2-8}
 & DDCD & 2.0 & 14.7 & 96.0 & 86.4 & 85.8 & 83.7 \\
\hline
\end{tabular}
\vspace{-0.2cm}
\end{table}

Table \ref{Table_Comparison_code_gadgets} summarizes the results by setting FPR to 2.0\% in each model because 2.0\% is 
%selected because it is
the FPR of the model achieving the highest F1-measure (i.e., BGRU using data dependency and control dependency).
For models learned from the datasets that accommodate data dependency, we observe that CNN and bidirectional RNNs (i.e., BLSTM and BGRU) are much more effective than DBN and shallow learning models (i.e., LR and MLP).
When compared with the models that are learned from the datasets that accommodate data dependency only, we observe that the models learned from the datasets that accommodate both data dependency and control dependency can improve
%the vulnerability detection capability in almost every scenario,
 FNR by 30.4\% and F1-measure by 24.0\% on average.
This can be explained by the fact that control dependency accommodates extra information useful to vulnerability detection.
%distinguish vulnerable code from the code that is not vulnerable.

\begin{insight}
\em A model which accommodates more semantic information (i.e., control dependency and data dependency) achieves a higher vulnerability detection capability.
\end{insight}

\subsubsection{Experiments for Answering RQ4}
\label{Experiment_RQ4}

We consider BGRU learned from the 341,536 SeVCs corresponding to the 4 kinds of SyVCs extracted from the training programs and the 79,091
%111,733
SeVCs extracted from the test programs, while accommodating semantic information induced by data dependency and control dependency.
We compare our most effective model BGRU with the commercial static vulnerability detection tool Checkmarx \cite{Checkmarx} and open-source static analysis tools Flawfinder \cite{FlawFinder} and RATS \cite{RATS},
%because they are available to us and widely used.
because (i) these tools arguably represent the state-of-the-art static analysis for vulnerability detection; (ii) they are widely used for detecting vulnerabilities in C/C++ source code; (iii) they directly operate on the source code (i.e., no need to compile the source code); and (iv) they are available to us.
We also consider the state-of-the-art system VUDDY \cite{kim2017vuddy}, which is particularly suitable for detecting vulnerabilities incurred by code cloning.
We further consider VulDeePecker \cite{vuldeepecker}, and we consider all 4 kinds of SyVCs and data as well as control dependency for SySeVR.
%which used the BLSTM trained from FC-kind SyVCs while accommodating semantic information induced by data dependency only.

\begin{table}[!htbp]
\vspace{-0.2cm}
\caption{\small Comparing BGRU in the SySeVR framework and state-of-the-art vulnerability detectors (metrics unit: \%)}
\vspace{-0.2cm}
\label{Table_Comparison_with_other_tools}
\centering
\footnotesize
\begin{tabular}{|c|c|c|c|c|c|c|c|}
\hline
Method & \tabincell{c}{FPR} &  \tabincell{c}{FNR} &  \tabincell{c}{A}&  \tabincell{c}{P} &  \tabincell{c}{F1} & MCC\\
\hline
Flawfinder & 21.6 & 70.4 & 69.8 & 22.8 & 25.7 & 22.1\\
\hline
RATS  & 21.5 & 85.3 & 67.2 & 12.8 & 13.7 & 12.6\\
\hline
Checkmarx & 20.8 & 56.8 & 72.9 & 30.9 & 36.1 & 33.0\\
\hline
VUDDY & 4.3 & 90.1 & 71.2 & 47.7 & 16.4 & 15.2\\
\hline
VulDeePecker & 2.5 & 41.8 & 92.2 & 78.0 & 66.6 & 64.9\\
\hline
SySeVR-BGRU & 1.4 & 5.6 & 98.0 & 90.8 & 92.6 & 90.5\\
\hline
\end{tabular}
\vspace{-0.2cm}
\end{table}

\begin{table*}[!htbp]
\vspace{-0.2cm}
\caption{\small The 15 vulnerabilities, which are detected by BGRU but not reported in the NVD, include 7 unknown vulnerabilities and 8 vulnerabilities that have been ``silently'' patched.}
\vspace{-0.2cm}
\label{Table_vulnerabilities detected}
\centering
\footnotesize
\begin{tabular}{|c|c|c|c|c|c|c|}
\hline
Target product & CVE ID & \tabincell{c}{Vulnerable\\ product reported} & \tabincell{c}{Vulnerability \\release date} & \tabincell{c}{Vulnerable file in\\ the target product} & \tabincell{c}{Kind\\ of SyVC}  & \tabincell{c}{1st patched version \\of target product} \\
\hline
Libav 10.3 & CVE-2013-7020 & FFmpeg & 12/09/2013 & libavcodec/ffv1dec.c & PU-kind & Libav 10.4 \\
\hline
{\multirow{4}{*}{\tabincell{c}{Libav 10.3,\\ Libav 12.3}}} & CVE-2013-**** & FFmpeg & **/**/2013 &libavcodec/**.c & AU-kind & -- \\
\cline{2-7}
 & CVE-2013-**** & FFmpeg & **/**/2013 & libavcodec/**.c & PU-kind & -- \\
\cline{2-7}
 & CVE-2014-**** & FFmpeg & **/**/2015 & libavcodec/**.c & PU-kind & -- \\
\cline{2-7}
 & CVE-2014-**** & FFmpeg & **/**/2014 & libavcodec/**.c & PU-kind & -- \\
\hline
Libav 9.10 & CVE-2014-9676 & FFmpeg & 02/27/2015 & libavformat/segment.c & PU-kind & Libav 10.0 \\
\hline
{Seamonkey 2.32} & CVE-2015-4511 & Firefox & 09/24/2015 & .../src/nestegg.c & AU-kind & Seamonkey 2.38\\
\hline
{Seamonkey 2.35} & CVE-2015-**** & Firefox & **/**/2015  & .../gonk/**.cpp & FC-kind & -- \\
\hline
 {\multirow{2}{*}{Thunderbird 38.0.1}} & CVE-2015-4511 & Firefox & 09/24/2015 & .../src/nestegg.c & AU-kind & Thunderbird 43.0b1\\
\cline{2-7}
& CVE-2015-**** & Firefox & **/**/2015  & .../gonk/**.cpp & FC-kind & -- \\
\hline
 {\multirow{3}{*}{Xen 4.4.2}} & CVE-2013-4149 & Qemu & 11/04/2014 & .../net/virtio-net.c & PU-kind & Xen 4.4.3\\
\cline{2-7}
 & CVE-2015-1779 & Qemu & 01/12/2016 & ui/vnc-ws.c & PU-kind & Xen 4.5.5 \\
\cline{2-7}
& CVE-2015-3456 & Qemu & 05/13/2015 & .../block/fdc.c & PU-kind & Xen 4.5.1 \\
%& CVE-2015-3456 & Qemu &  & hw/block/fdc.c & PU & Xen 4.5.1 \\
\hline
{Xen 4.7.4} & CVE-2016-4453 & Qemu & 06/01/2016 & .../display/vmware\_vga.c & AE-kind & Xen 4.8.0\\
\hline
\tabincell{c}{Xen 4.8.2,\\ Xen 4.12.0} & CVE-2016-**** & Qemu & **/**/2016 & .../net/**.c & PU-kind & -- \\
\hline
\end{tabular}
\vspace{-0.2cm}
\end{table*}

Table \ref{Table_Comparison_with_other_tools} summarizes the experimental results.
We observe that SySeVR-enabled BGRU substantially outperforms the state-of-the-art vulnerability detection methods.
The open-source Flawfinder and RATS
%, the simple parsers and imperfect patterns make the
have high FPRs and FNRs.
% of these tools prohibitively high. %\cite{DBLP:phd/dnb/Yamaguchi15}.
Checkmarx is better than Flawfinder and RATS, but still has
%can provide high parsing capabilities. However, the
high FPRs and FNRs.
VUDDY is known to trade a high FNR for a low FPR,
%, which leads to a very low F1-measure.
because it can only detect vulnerabilities that are nearly identical to the vulnerabilities in the training programs.
SySeVR-enabled BGRU is much more effective than VulDeePecker because VulDeePecker cannot cope with other kinds of SyVCs (than FC) and cannot accommodate semantic information induced by control dependency.
Moreover, BGRU learned from a larger training set (i.e., 341,536 SeVCs) is more effective than BGRU learned from a smaller training set (30,000 SeVCs; see Table \ref{Table_neural_networks_four_types_SyVCs_FPR0.2}), especially reducing FNR by 9.1\%.
%by reducing 9.1\% in FNR. 
In summary,

\begin{insight}
{\em SySeVR-enabled BGRU is much more effective than the state-of-the-art vulnerability detection methods.}
\end{insight}

\subsubsection{Applying BGRU to Detect Vulnerabilities in Software Products}

In order to show the usefulness of SySeVR in detecting software vulnerabilities in real-world software products,
we apply SySeVR-BGRU trained in Section \ref{Experiment_RQ4} to detect vulnerabilities in 4 software products: Libav, Seamonkey, Thunderbird, and Xen.
Each of these products contains multiple targets programs, from which we extract their SyVCs, SeVCs, and vectors.
For each product, we apply SySeVR-enabled BGRU to its 20 versions so that we can tell whether some vulnerabilities have been ``silently'' patched by the vendors when releasing a newer version.

As highlighted in Table \ref{Table_vulnerabilities detected}, we detect 15 vulnerabilities that are {\em not} reported in NVD.
Among them, 7 are unknown (i.e., %0-day in the sense that
their presence in these products are not known until now) and are indeed similar (upon our manual examination) to the {\em CVE IDentifiers} (CVE IDs) mentioned in Table \ref{Table_vulnerabilities detected}. We do not give the full details of these vulnerabilities %in this paper
for ethical considerations, but we have
reported these 7 vulnerabilities to the vendors.
The other 8 vulnerabilities have been ``silently'' patched by the vendors when releasing newer versions of the products in question.
Checkmarx, which we use to extract vulnerability syntax characteristics, missed all of these vulnerabilities except the two in Seamonkey 2.35 and Thunderbird 38.0.1, demonstrating its ineffectiveness.

\section{Limitations}
\label{sec:Limitations}
The present study has several limitations.
{\bf First}, we focus on detecting vulnerabilities in C/C++ program source code,
meaning that the framework may need to be adapted to cope with other programming languages and/or executables.
{\bf Second}, our experiments focus 4 kinds of vulnerability syntax characteristics, which cover 93.6\% of the vulnerable programs collected from SARD. This coverage is not perfect while noting that the SARD data may not be representative of real-world software products. Future research needs to identify more complete vulnerability syntax characteristics.
{\bf Third},
the algorithms for generating SyVCs and SeVCs could be improved to accommodate more syntactic/semantic information for vulnerability detection.
{\bf Fourth}, our experiments use a single model to detect multiple kinds of vulnerabilities. Future research should investigate which of the following is more effective: using multiple models that are respectively tailored to detect multiple kinds of vulnerabilities vs. using a single model to detect multiple kinds of vulnerabilities.
{\bf Fifth}, we detect vulnerabilities at the slice level (i.e., multiple lines of code that are semantically related to each other), which
could be improved to more precisely pin down the {\em line} of code that contains a vulnerability.
{\bf Sixth}, we generate ground-truth labels by manually checking 0.6\% of all samples, which may have been mislabeled by the automatic method we use (owing to the lack of ground-truth dataset). 
Future research should investigate more effective automatic labeling methods;
for this purpose, one may leverage the idea of co-training \cite{DBLP:journals/ngc/CaldasGM18}.
{\bf Seventh}, our experiments show some deep neural networks are more effective than the state-of-the-art vulnerability detection methods.
Although we have gained some insights into explaining the ``why'' part, more investigations are needed to explain the success of deep learning in this context and beyond.

\section{Related Work}
\label{sec:Related_work}
\noindent{\bf Prior studies related to vulnerability detection}.
There are two methods for source code-based static vulnerability detection: {\em code similarity-based} vs. {\em pattern-based}.
Since code similarity-based detectors can only detect vulnerabilities incurred by code cloning
and SySeVR is a pattern-based method,
we only review prior studies in pattern-based methods,
which can be further divided into {\em rule-based} and {\em machine learning-based} methods.

Rule-based methods use vulnerability patterns to detect vulnerabilities, where patterns are manually generated by human experts
(e.g., Flawfinder \cite{FlawFinder}, RATS \cite{RATS}, Checkmarx \cite{Checkmarx}).
These tools often incur high false-positive rates and/or high false-negative rates \cite{DBLP:phd/dnb/Yamaguchi15}, as also confirmed by our experiments (Section \ref{Experiment_RQ4}). Vulnerability patterns can be defined using, for example, {\em code property graphs} \cite{yamaguchi2014modeling}.
In contrast, SySeVR uses vulnerability patterns that are learned automatically and represented by deep neural networks.

Machine learning-based methods, as discussed elsewhere  \cite{DBLP:journals/csur/GhaffarianS17},
can be further divided into the following three sub-categories.
(i) Vulnerability prediction methods based on software metrics: These methods are built on top of software metrics
% that are ofdefined manually,
(e.g., imports and function calls \cite{neuhaus2007predicting}, complexity \cite{DBLP:conf/issre/WaldenSS14,shin2011evaluating}, code churn and developer activity \cite{shin2011evaluating,DBLP:conf/esem/MeneelySMTMS13}), but predict vulnerabilities at a coarse granularity (e.g., component-level \cite{neuhaus2007predicting} or file-level \cite{shin2011evaluating}), meaning that they cannot pin down the locations of vulnerabilities.
(ii) Anomaly detection methods: These methods find vulnerabilities via abnormal patterns in
%that do not conform to the normal or expected ones 
(for example) API usage \cite{DBLP:conf/issta/GruskaWZ10} or missing checks \cite{yamaguchi2013chucky,DBLP:journals/tse/ChangPY08}, but cannot cope with rarely-used but normal patterns.
(iii) Vulnerable code pattern recognition methods:
These methods extract vulnerability patterns related to (for example) ASTs %(Abstract Syntax Trees) 
\cite{yamaguchi2012generalized}, code property graphs \cite{yamaguchi2015automatic} or system calls \cite{grieco2016toward}, and use these patterns to detect vulnerabilities.
These methods demand human experts to define features and use the traditional machine learning models (e.g., support vector machine and $k$-nearest neighbor) to detect vulnerabilities.
Recently, deep learning has been leveraged for vulnerability detection, while alleviating the problem of manual feature definition.
Lin et al. \cite{DBLP:conf/ccs/LinZLPX17} presented a method for automatically learning high-level representations of functions (i.e., coarse-grained).
% and not able to pin down locations of vulnerabilities).
VulDeePecker \cite{vuldeepecker} is the first system showing the feasibility of using deep learning to detect vulnerabilities at the slice level, which is much finer than the function level.
A more recent development is $\mu$VulDeePecker \cite{muVulDeePecker}, which extends VulDeePecker to detect {\em multiclass} vulnerabilities.
SySeVR overcomes the weaknesses of VulDeePecker discussed in Section \ref{sec:Introduction}, and is the first systematic framework for using deep learning to detect vulnerabilities.

\noindent{\bf Prior studies related to deep learning}. Deep learning has been used for program analysis.
CNN has been used for software defect prediction \cite{DBLP:conf/qrs/LiHZL17} and locating buggy source code \cite{DBLP:conf/ijcai/HuoLZ16};
DBN has been used for software defect prediction \cite{wang2016automatically, DBLP:conf/qrs/YangLXZS15};
RNN has been used for vulnerability detection \cite{vuldeepecker, DBLP:conf/ccs/LinZLPX17, muVulDeePecker},
software traceability \cite{DBLP:conf/icse/0004CC17}, code clone detection \cite{white2016deep}, and recognizing functions in binaries \cite{shin2015recognizing}.
The present study offers the first framework for using deep learning to detect vulnerabilities.

\section{Conclusion}
\label{sec:Conclusion}

We presented the SySeVR framework for using deep learning to detect vulnerabilities.
Based on a large dataset of vulnerability we collected, we drew a number of insights, including an explanation on the effectiveness of deep learning in vulnerability detection.
Moreover, we detected
%, from 4 software products,
15 vulnerabilities that were {\em not} reported in the NVD.
Among these 15 vulnerabilities, 7 are unknown %(i.e., 0-day)
and have been reported to the vendors, and the other 8 have been ``silently'' patched by the vendors when releasing newer versions.
There are many open problems for future research. In addition to addressing the limitations discussed in Section \ref{sec:Limitations},
it is important to investigate the impact of {\em code duplication} \cite{CodeDuplication} on SySeVR-enabled models.

\section*{Acknowledgment}
We thank the reviewers for their constructive comments, which have guided us in improving the paper.
We thank Sujuan Wang and Jialai Wang for collecting the vulnerable programs from NVD and SARD. The authors from Huazhong University of Science and Technology and Hebei University were supported in part by the National Natural Science Foundation of China under Grant No. U1936211 and No. 61802106 and in part by the Natural Science Foundation of Hebei Province under Grant No. F2020201016. 
S. Xu was supported in part by ARO Grant \#W911NF-17-1-0566 as well as NSF Grants \#1814825 and \#1736209. Any opinions, findings, conclusions or recommendations expressed in this work are those of the authors and do not reflect the views of the funding agencies in any sense.

%\appendices
%% you can choose not to have a title for an appendix
%% if you want by leaving the argument blank
%\section{}

%\ifCLASSOPTIONcompsoc
%  % The Computer Society usually uses the plural form
%\else
%  % regular IEEE prefers the singular form
%\fi

\ifCLASSOPTIONcaptionsoff
  \newpage
\fi



\bibliographystyle{IEEEtran}
\bibliography{bibliography}

\end{document}